\documentclass[letterpaper,twocolumn,10pt]{article}
\usepackage{usenix-2020-09}

\usepackage{graphicx}
\usepackage{amsmath}
\usepackage{amssymb}
\usepackage{booktabs}

\usepackage{soul}
\usepackage{times}
\usepackage{epsfig}
\usepackage{bbm}
\usepackage{mathrsfs}
\usepackage{multirow}
\usepackage{paralist}
\usepackage{mathtools}
\usepackage{url}            
\usepackage{booktabs}       
\usepackage{nicefrac}       
\usepackage{threeparttable}
\usepackage{color}

\usepackage{listings}
\usepackage{wrapfig}

\usepackage{pifont}
\usepackage[utf8]{inputenc}
\usepackage{comment}
\usepackage{amsmath,amssymb,amsfonts}
\usepackage{makecell}
\usepackage{indentfirst}
\usepackage{shorttoc}
\usepackage{caption}

\usepackage{cite}
\usepackage{algorithmic}
\usepackage{graphicx}
\usepackage{textcomp}
\usepackage{tabularx} 
\usepackage{collcell}

\usepackage{array}
\usepackage{caption}
\usepackage{hhline}
\usepackage[linesnumbered, ruled, vlined]{algorithm2e}
\usepackage[table,xcdraw]{xcolor}
\usepackage{diagbox}

\usepackage{array}
\newcolumntype{P}[1]{>{\centering\arraybackslash}p{#1}}

\usepackage{tikz}

\def\BibTeX{{\rm B\kern-.05em{\sc i\kern-.025em b}\kern-.08em
    T\kern-.1667em\lower.7ex\hbox{E}\kern-.125emX}}
\definecolor{deepred}{rgb}{0.631,0.102,0.102}
\definecolor{skyblue}{HTML}{126da2}
\definecolor{mildyellow}{HTML}{FFF2CC}
\definecolor{minired}{HTML}{000000}

\newcommand{\AlgName}{\textsc{ASSET}}

\newcommand{\yi}[1]{\textbf{\textcolor{red}{[Yi: #1]}}}
\newcommand{\minz}[1]{\textbf{\textcolor{purple}{[Minzhou: #1]}}}

\usepackage{authblk}

\makeatletter
\newcommand{\printfnsymbol}[1]{%
  \textsuperscript{\@fnsymbol{#1}}%
}
\makeatother

\begin{document}

\title{\AlgName: Robust Backdoor Data Detection Across a Multiplicity of Deep Learning Paradigms}

\author[1]{Minzhou Pan\thanks{ \textbf{Y. Zeng} and \textbf{M. Pan} contributed equally. Correspond \href{mailto:yizeng@vt.edu}{Y. Zeng} or \href{mailto:ruoxijia@vt.edu}{R. Jia.}}\printfnsymbol{1}}
\author[1]{Yi Zeng\printfnsymbol{1}}
\author[2]{Lingjuan Lyu}
\author[3]{Xue Lin}
\author[1]{Ruoxi Jia}

\affil[1]{Virginia Tech, Blacksburg, VA 24061, USA}
\affil[2]{Sony AI, Tokyo, 108-0075, Japan}
\affil[3]{Northeastern University, Boston, MA 02115, USA}

\renewcommand\Authands{ and }

\maketitle


\begin{abstract}
\vspace{-.5em}

Backdoor data detection is traditionally studied in an end-to-end supervised learning (SL) setting. However, recent years have seen the proliferating adoption of self-supervised learning (SSL) and transfer learning (TL), due to their lesser need for labeled data. Successful backdoor attacks have also been demonstrated in these new settings. However, we lack a thorough understanding of the applicability of existing detection methods across a variety of learning settings. 
By evaluating 56 attack settings, we show that the performance of most existing detection methods varies significantly across different attacks and poison ratios, and all fail on the state-of-the-art clean-label backdoor attack which only manipulates a few training data's features with imperceptible noise without changing labels. In addition, existing methods either become inapplicable or suffer large performance losses when applied to SSL and TL. We propose a new detection method called \underline{A}ctive \underline{S}eparation via Off\underline{set} ($\AlgName$), which actively induces different model behaviors between the backdoor and clean samples to promote their separation. We also provide procedures to adaptively select the number of suspicious points to remove. In the end-to-end SL setting, $\AlgName$ is superior to existing methods in terms of consistency of defensive performance across different attacks and robustness to changes in poison ratios; in particular, it is the only method that can detect the state-of-the-art clean-label attack. Moreover, $\AlgName$'s average detection rates are higher than the best existing methods in SSL and TL, respectively, by 69.3\% and 33.2\%, thus providing the first practical backdoor defense for these emerging DL settings. 

\end{abstract}

\vspace{-1.2em}
\section{Introduction}
\vspace{-1em}

\begin{figure}[t!]
  \centering
     \vspace{-1.5em}
  \includegraphics[width=0.85\linewidth]{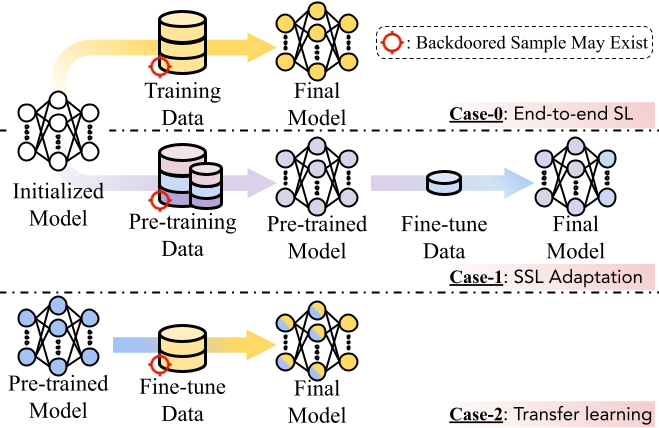}
  \vspace{-1.em}
  \caption{Illustration of popular DL paradigms and corresponding threat models. \underline{\textbf{Case-0}}: traditional end-to-end SL, where one trains a model from scratch.  \underline{\textbf{Case-1}}: SSL adaptation, where one first pre-trains a model via SSL using unlabeled pre-training data and then linearly adapts to a small amount of labeled data to obtain the final model.  \underline{\textbf{Case-2}}: TL, where one starts with an existing pre-trained model and fine-tunes it.
Existing work has demonstrated successful attacks in all three cases under the threat models where datasets marked with \textcolor{red}{red} circle indicates poisons.
Yet, none of the existing backdoor detection methods is evaluated in all three cases.
  }
  \label{fig:compare}
  \vspace{-1.5em}
\end{figure}

Deployment of deep learning (DL) in critical services and infrastructures calls for special emphasis on security, given its susceptibility to erroneous predictions in the presence of attacks \cite{gu2017badnets, zeng2022narcissus,szegedy2013intriguing}. Specifically, data-poisoning-based \textit{backdoor attacks} - where attackers manipulate the training data to force certain outputs during testing - pose a significant threat. Successful attacks have been demonstrated on various computer vision tasks and beyond~\cite{li2020backdoor}. This paper focuses on the problem of \emph{detecting the poisoned samples within a training set}. An effective detection strategy allows one to mitigate the risk of backdoors by removing suspicious samples from training.

Poisoned samples can be regarded as outliers in a training set. However, unlike \emph{arbitrary} outliers considered in the classical outlier detection and robust statistics literature, poisoned samples are special outliers that induce \emph{specific model behaviors}, e.g., misleading the model to predict some target class(es). 
Hence, recent works on backdoor detection primarily leverage the model trained on the poisoned dataset (backdoored model hereinafter ) or information cached during training to help discover poisoned samples \cite{tang2021demon,gao2019strip,tran2018spectral,hayase2021spectre,ma2022beatrix,chen2018detecting,li2021anti}. For instance, most of the prior work starts by extracting the backdoored model's output~\cite{gao2019strip}, intermediate activation patterns~\cite{tran2018spectral,hayase2021spectre,ma2022beatrix,chen2018detecting}, gradient~\cite{li2021anti} for each sample, and then separate poisons from clean samples based on the extracted information. 

While taking advantage of the information collected from the downstream learning process provides a clear path to enhancing backdoor detection performance, it also raises the question: 
\emph{Can these detection methods maintain their performance across different DL settings?}
Particularly, existing detection methods are exclusively evaluated in only one learning setting---\emph{end-to-end supervised learning} (SL), where a labeled poisoned dataset is used to train a model \emph{from scratch}. On the other hand, new learning paradigms are increasingly adopted and have demonstrated state-of-the-art prediction performance with reduced annotation costs and computational burden~\cite{chen2020simple,he2022masked,dosovitskiy2020image,haochen2022beyond}. 
The two most representative and popular paradigms are \emph{self-supervised learning (SSL) adaptation} and \emph{transfer learning (TL)}, as illustrated in Figure~\ref{fig:compare}. 

In SSL adaptation, one pre-trains a model on large \emph{unlabeled} data (e.g., through contrastive learning~\cite{chen2020big,chen2021empirical,grill2020bootstrap} or masked autoencoder (MAE)~\cite{he2022masked}) and then fine-tunes only the last layer using \emph{labeled} data from a specific downstream task. Recent work~\cite{carlini2021poisoning,saha2022backdoor,li2022demystifying} has shown that an attacker can poison the unlabeled dataset to implant backdoors without any control over downstream fine-tuning processes. 
Thus, it is natural to ask: \emph{Can we detect the poisoned samples within an unlabeled dataset using existing methods?}
%
In TL, one starts with an existing pre-trained model and fine-tunes all layers of the model or just the last layer with labeled data. 
Despite the importance of TL in practice~\cite{raffel2020exploring}, we lack an understanding of backdoor detection in this setting:
\emph{Can we detect the poisoned samples when they are used for fine-tuning an existing model instead of training it from scratch?} 

\textbf{Our first contribution is a comprehensive evaluation of existing detection methods across different DL paradigms.} The key findings are summarized as follows.
\begin{compactitem}
    \item \textbf{(\underline{Case-0}) End-to-end SL:} 
    Despite the efficiency demonstrated by prior detection efforts in specific settings, the consistency of efficacy varies a lot across different attacks or poison ratios.
    In particular, all fail to detect the state-of-the-art clean-label backdoor attack\footnote{Clean-label attacks refer to those where the poisoned samples appear to be correctly labeled to a human inspector.}~\cite{zeng2022narcissus} and underperform in the very low or very high poison ratio setting (e.g., 0.05\% or 20\%). 
    
    \item \textbf{(\underline{Case-1}) SSL adaptation:} 
    There are no existing methods dedicated to detecting unlabeled poisoned samples in the SSL setting. Yet, some of the existing methods can be adapted to the SSL. For instance, those methods attempting to separate the poisoned samples from clean in the embedding space can employ an embedder learned from unlabeled data to generate the embedding for each sample~\footnote{We will elaborate on the adaptation techniques in Section~\ref{sec:settings}.}. However, the performance of these methods after adaptation is limited (e.g., their average detection rate over different attacks all falls below 26\%).


    \item \textbf{\underline{Case-2} TL:} While prior literature omitted TL in their evaluation, the detection methods can all be applied to it. However, the methods based on embeddings suffer a significant performance loss compared to the end-to-end SL setting because the poisoned samples are less distinguishable from clean ones in a fine-tuned embedding space than a trained-from-scratch one.
\end{compactitem}

The limitations of existing methods per our evaluation are summarized in Table~\ref{tab:compare}. Overall, there still lacks a detection method that is effective across different learning paradigms.

\renewcommand{\arraystretch}{1.3}
\begin{table}[t!]
\centering
\vspace{-1.5em}
\resizebox{\columnwidth}{!}{
\begin{tabular}{p{2.3cm}<{\centering}| p{1.3cm}<{\centering}| p{1.3cm}<{\centering}| p{1.3cm}<{\centering}| p{1.3cm}<{\centering}| p{1.3cm}<{\centering}| p{1.3cm}<{\centering}| p{1.3cm}<{\centering}| p{1.3cm}<{\centering}}
\hline
 &
 
\textbf{\begin{tabular}[c]{@{}c@{}}Spectral \\ \cite{tran2018spectral}\end{tabular}} &
\textbf{\begin{tabular}[c]{@{}c@{}}Spectre \\ \cite{hayase2021spectre}\end{tabular}} &
\textbf{\begin{tabular}[c]{@{}c@{}}Beatrix \\ \cite{ma2022beatrix}\end{tabular}} &
\textbf{\begin{tabular}[c]{@{}c@{}}AC \\ \cite{chen2018detecting}\end{tabular}} &
\textbf{\begin{tabular}[c]{@{}c@{}}ABL \\ \cite{li2021anti}\end{tabular}} &
\textbf{\begin{tabular}[c]{@{}c@{}}Strip \\ \cite{gao2019strip}\end{tabular}} &
\textbf{\begin{tabular}[c]{@{}c@{}}CT \\ \cite{qi2022fight}\end{tabular}} &
  \textbf{Ours} \\ \hline
  
  \textbf{\begin{tabular}[c]{@{}c@{}}Applicable to \\ Labeled Data\end{tabular}} &
  \cellcolor[HTML]{D9EAD3}$\checkmark$ &
  \cellcolor[HTML]{D9EAD3}$\checkmark$ &
  \cellcolor[HTML]{D9EAD3}$\checkmark$ &
  \cellcolor[HTML]{D9EAD3}$\checkmark$ &
  \cellcolor[HTML]{D9EAD3}$\checkmark$ &
  \cellcolor[HTML]{D9EAD3}$\checkmark$ &
  \cellcolor[HTML]{D9EAD3}$\checkmark$ &
  \cellcolor[HTML]{D9EAD3}$\checkmark$ \\ \hline
\textbf{\begin{tabular}[c]{@{}c@{}}Applicable to \\ Unlabeled Data\end{tabular}} &
  \cellcolor[HTML]{FFF2CC}$\bigcirc$ &
  \cellcolor[HTML]{FFF2CC}$\bigcirc$ &
  \cellcolor[HTML]{FFF2CC}$\bigcirc$ &
  \cellcolor[HTML]{FFF2CC}$\bigcirc$ &
  \cellcolor[HTML]{FFF2CC}$\bigcirc$ &
  \cellcolor[HTML]{FFF2CC}$\bigcirc$ &
  \cellcolor[HTML]{F4CCCC}$\times$ &
  \cellcolor[HTML]{D9EAD3}$\checkmark$ \\ \hline
\textbf{\begin{tabular}[c]{@{}c@{}}Robust to \\ Diff. Triggers \end{tabular}} &
  \cellcolor[HTML]{F4CCCC}$\times$ &
  \cellcolor[HTML]{F4CCCC}$\times$ &
  \cellcolor[HTML]{F4CCCC}$\times$ &
  \cellcolor[HTML]{F4CCCC}$\times$ &
  \cellcolor[HTML]{F4CCCC}$\times$ &
  \cellcolor[HTML]{F4CCCC}$\times$ &
  \cellcolor[HTML]{F4CCCC}$\times$ &
  \cellcolor[HTML]{D9EAD3}$\checkmark$ \\ \hline
\textbf{\begin{tabular}[c]{@{}c@{}}Robust to Diff. \\  Poison Ratios \end{tabular}} &
  \cellcolor[HTML]{F4CCCC}$\times$ &
  \cellcolor[HTML]{F4CCCC}$\times$ &
  \cellcolor[HTML]{F4CCCC}$\times$ &
  \cellcolor[HTML]{F4CCCC}$\times$ &
  \cellcolor[HTML]{F4CCCC}$\times$ &
  \cellcolor[HTML]{F4CCCC}$\times$ &
  \cellcolor[HTML]{F4CCCC}$\times$ &
  \cellcolor[HTML]{D9EAD3}$\checkmark$ \\ \hline
\end{tabular}}
\vspace{-1.em}
\caption{A summary and comparison of representative works in the detection of backdoored samples.
\scalebox{0.55}{\colorbox[HTML]{FFF2CC}{$\bigcirc$}}
denotes partially satisfactory (i.e., requiring additional adaptation).
} 
\label{tab:compare}
\vspace{-1.5em}
\end{table}

\textbf{Our second contribution is the development of a robust, generic approach to backdoor detection that applies to the three representative learning paradigms discussed above.} Like most existing literature~\cite{ma2022beatrix, gao2019strip, hayase2021spectre}, our approach also assumes that the defender has an extra set of clean samples (referred to as a \emph{base set} hereinafter) with a size much smaller compared to the training set. In practice, these clean samples can be obtained through manual inspection or automatic screening~\cite{zeng2022sift}. However, unlike the previous works, we do not require the base set to be labeled. 

The key idea of our approach is to induce different model behaviors between poisoned samples and clean ones. To achieve this, we design a two-step optimization process: we first \emph{minimize} some loss on the clean base set; then, we attempt to \emph{offset} the effect of the first minimization on the clean distribution by \emph{maximizing} the same loss on the entire training set including both clean and poisoned samples. The outcome of this two-step process is a model which returns high loss for poisoned samples and low loss for clean ones. Hence, we can decide whether a sample is poisoned or clean based on the corresponding loss value.
%

We found that the two-step optimization-based offset idea achieves strong detection performance except in settings where the poison ratio is low, or the learning of the poisoned samples happens slowly---at roughly the same speed as learning of clean samples. As we will explicate later in the paper, in these cases, the effect of the second maximization significantly outweighs that of the first minimization; as a result, both poisoned and clean samples achieve large losses and become inseparable. 

To tackle the challenge, we propose a strengthened technique that involves \emph{two nested offset procedures}, and the inner offset reinforces the outer one. Specifically, we use the inner offset procedure to identify the points most likely to be poisoned and mark them as suspicious; the outer offset procedure still minimizes some loss on the clean base set, but the maximization will now be performed on the points marked to be suspicious by the inner offset, instead of the entire poisoned dataset.
As the proportion of clean samples within the suspicious set is much smaller than that within the entire poisoned set, the small loss of clean samples obtained from the first minimization would be impacted much less by the second maximization. 
This nested design effectively improves the separability between clean and poisoned samples.

\textbf{Our third contribution is the provision of techniques that can adaptively set the loss threshold to discern poisoned samples.} Some of the prior works~\cite{tran2018spectral,hayase2021spectre} assume the knowledge of poison ratio and mark a fixed number of samples as poisoned ones based on their respective criteria. Moreover, the poisoned and clean samples often do not have a clear separation based on their criteria (see examples in Figure \ref{fig:slcompare}); as a result, their detection performance is very sensitive to the estimated poison ratio. We argue that in practice, it is challenging to have an accurate estimate of the poison ratio. Hence, it is preferable to adapt detection to the data characteristics rather than relying on a fixed estimate. Herein, we design two adaptive thresholding techniques tailored to specific requirements imposed by inner and outer offset procedures (i.e., prioritizing 
precision vs. prioritizing true positive rate).  
%

We conduct extensive experiments in comparison with seven representative or state-of-art backdoor data detection methods over 56 different attack settings across various DL paradigms and show that our proposed method, $\AlgName$, is the only one that can provide reliable detection consistently across all the evaluated settings. This work is also the first practical backdoor detection for the SSL and the TL settings\footnote{Open-source: \url{https://github.com/ruoxi-jia-group/ASSET}}.

\vspace{-1.em}
\section{Background \& Related Work}
\vspace{-.5em}
\label{sec:background}

\noindent\textbf{End-to-end supervised learning \& transfer learning.}
The objective of end-to-end SL is to train a classifier $f(\cdot|\theta):\mathcal{X} \rightarrow[k]$, which predicts the label $y \in [k]$ of an input $x\in \mathcal{X}$. $\theta$ denotes the parameters of the classifier $f(\cdot|\theta)$. The standard end-to-end SL (\textbf{\underline{Case-0}}) consists of two stages: training and testing. In the training stage, a learning algorithm is provided with a set of training data, $D=\{(x_i,y_i)\}_{i=1}^N$, consisting of examples from $k$ classes. Then, the learning algorithm seeks the model parameters, $\theta$, that minimize the empirical risk:
\vspace{-1.em}
\begin{equation}
\begin{aligned}
\theta^{*} = \arg \min _{\theta} \sum_{i=1}^N \mathcal{L}\left(f\left(x_i|\theta\right), y_i\right).
\label{equ:learning}
\end{aligned}
\vspace{-1.em}
\end{equation}
When $f(\cdot|\theta)$ is a deep neural network, the corresponding empirical risk is a non-convex function of $\theta$, and finding a global minimum is generally impossible. Hence, the standard practice is to look for a local minimum. Algorithmically, the model is initialized with random parameters and updated iteratively via stochastic gradient descent~\cite{bottou2012stochastic}. In the test stage, the trained model $f(\cdot|{\theta^*})$ takes input test examples and serves up predictions. 
TL (\textbf{\underline{Case-2}}) shares the same optimization goal as the end-to-end SL. However, TL initializes the optimization with a pre-trained backbone model instead of random parameters. Within the scope of this paper, we consider two of the most popular TL schemes: (1) \textsf{FT-all}: the entire pre-trained model gets updated during training (e.g., \cite{dosovitskiy2020image,he2022masked,tan2019efficientnet}); (2) \textsf{FT-last} (or linear adaptation): only the last fully-connected layer is updated (e.g., \cite{chen2020big,xie2021self,haochen2022beyond}). In the context of TL, we will refer to solving the optimization (\ref{equ:learning}) as \emph{fine-tuning} and $D$ as the \emph{fine-tuning data}.

\begin{figure}[t!]
\vspace{-1.5em}
  \centering
  \includegraphics[width=0.9\linewidth]{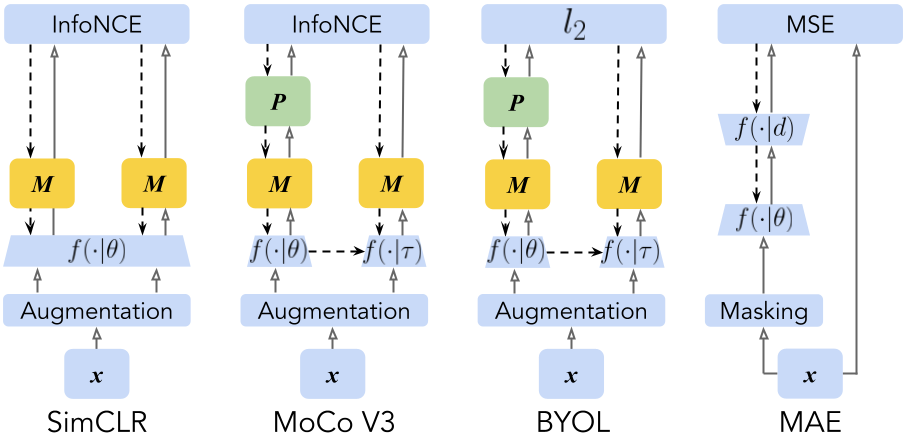}
  \vspace{-0.5em}
  \caption{
  Illustration of representative SSL methods: SimCLR \cite{chen2020simple}, MoCo V3 \cite{chen2020simple}, BYOL \cite{grill2020bootstrap}, and the Masked Auto-Encoder \cite{he2022masked}. The solid gray arrow indicates forward propagation, and the dashed black arrow indicates backpropagation.
  }
  \label{fig:contrastive}
  \vspace{-1.5em}
\end{figure}

\noindent \textbf{Self-supervised learning.}
SSL usually consists of two phases: pretext training and fine-tuning. Pretext training aims to train an encoder $f(\cdot|\theta):\mathcal{X} \rightarrow \mathcal{Z}$ that can map the input $x\in \mathcal{X}$ into the embedding $z\in \mathcal{Z}$. $\theta$ denotes the parameters of the encoder $f(\cdot|\theta)$. This paper focuses primarily on two of the most recent SSL schemes: contrastive learning and masked auto-encoder (MAE). Their training processes are illustrated in Figure \ref{fig:contrastive}, 
where $M$ is a multi-layer perceptron (MLP) used to reduce the dimension of features,
and $P$ is a predictor.
The fundamental idea of contrastive learning, e.g., SimCLR \cite{chen2020simple}, MoCo V3 \cite{chen2021empirical}, and BYOL \cite{grill2020bootstrap}, is to learn an encoder by bringing the embeddings corresponding to the augmentations of the same image (a.k.a. positive pairs) closer and distancing its embeddings from other images (a.k.a. negative pairs). 
All three methods pre-train $f(\cdot|\theta)$, $M$ and $P$ (if applicable) on large amounts of unlabeled data, and differ in how they generate positive and negative pairs and in the loss functions they use for training. We refer interested readers to~\cite{gui2023survey} for more details. 
By contrast, the recently proposed SSL method, MAE \cite{he2022masked}, trains the encoder $f(\cdot|\theta)$ by masking a portion of pixels in an image $x$ (the masked image is denoted by $x^{\prime}$) and then  using $f(x^{\prime}|\theta)$ with a decoder $d(\cdot)$ to restore $x$. 
For all the aforementioned SSL methods, after the pretext training, the acquired encoder parameters $\theta^*$ will be adapted to a downstream task similarly to TL using the fine-tuning data.

\noindent \textbf{Backdoor attacks.}
Backdoor attacks have been extensively studied in the end-to-end SL setting and can be categorized into dirty-label and clean-label attacks. Dirty-label backdoor attacks manipulate both label and feature of a sample. These attacks have developed from using a sample-independent visible pattern as the trigger \cite{gu2017badnets,chen2017targeted,liu2017trojaning,bagdasaryan2021blind} to more stealthy and powerful attacks with sample-specific \cite{li2021invisible} or visually imperceptible triggers \cite{li2020invisible,liu2020reflection,nguyen2021wanet,zeng2021rethinking,hammoud2021check,wang2022invisible}.
Clean-label backdoor attacks ensure that the manipulated features are semantically consistent with corresponding labels.
Existing attacks in this category range from inserting arbitrary triggers~\cite{turner2019label,saha2020hidden,souri2021sleeper} to optimized triggers~\cite{zeng2022narcissus}. Most of the above backdoor attacks can be easily adapted to TL settings without modifications. There are also backdoor attacks specifically designed for TL settings, e.g., the hidden trigger backdoor attack \cite{saha2020hidden}.

With the thriving development of SSL, especially contrastive learning (e.g., SimCLR \cite{chen2020simple,chen2020big}, MoCo \cite{he2020momentum,chen2020improved2,chen2021empirical}, BYOL \cite{grill2020bootstrap}) and the MAE \cite{he2022masked}, backdoor attacks targeting SSL have also been explored. Recent work mainly applies existing dirty-label backdoor triggers studied in SL to the targeted category of samples \cite{saha2022backdoor,carlini2021poisoning}.
However, attacks' efficacy are limited (ASR below 10\% on CIFAR-10 even with an in-class poison ratio set to 50\%, as shown in our experiment, Section \ref{sec:sslexp}). A recent attack \cite{li2022demystifying} exploits the ``representation invariance'' property of contrastive learning and instantiate a symmetric trigger via manipulation in the frequency domain, achieving much higher ASR with a lower poison ratio (e.g., in-class poison ratio of 10\%).

\noindent \textbf{Backdoor sample detection.}
Note that no existing backdoored sample detection methods have been considered nor evaluated over cases other than \underline{\textbf{Case-0}}. In particular, there is no practical defense under the SSL, and the study in TL is overlooked.
Many of the existing works identify poisoned samples by examining their difference from clean ones in the embedding space, such as using singular value decomposition (SVD) \cite{tran2018spectral,hayase2021spectre}, Gram matrix \cite{ma2022beatrix}, K-Nearest-Neighbors \cite{peri2020deep}, and
feature decomposition \cite{tang2021demon}.
In addition to embeddings, intermediate neural activation \cite{chen2018detecting,soremekun2020exposing} and gradients \cite{chan2019poison,chou2020sentinet} extracted from samples can also be adopted for backdoored sample detection.
Past work has also examined other differentiating properties of backdoor samples, such as trigger's resistance to augmentations \cite{gao2019strip}, high-frequency artifacts \cite{zeng2021rethinking}, low contribution to the training task \cite{wang2021unified,koh2017understanding}, or backdoor samples may achieve lower loss at the early stage of training \cite{li2021anti}.

A recent work \cite{qi2022fight} proposed a confusion training procedure, which trains a model on a weighted combination of the randomly-labeled clean base set and the poisoned set. Introducing a randomly-labeled clean set into training prevents the model from fitting to the clean portion of the poisoned data, thereby allowing the identification of poisoned samples whose labels are consistent throughout the training process. 
Our experiment found that the effectiveness of \cite{qi2022fight} highly relies on the hyperparameter tuning of the weighted combined-training process and the performance varies significantly with poison ratios. Additionally, the fundamental assumption is that decoupling the benign correlations between semantic features and semantic labels does not influence the learnability of the correlations between backdoor triggers and target labels. However, some advanced clean-label backdoor attack trigger~\cite{zeng2022narcissus} strongly entangles with the semantic features of the target class; therefore, \cite{qi2022fight} falls short of detecting the trigger.
At a high level, confusion training shares a similar idea to ours in the sense that we both leverage a clean base set to induce different detector behaviors between clean and poisoned samples. However, there are several key differences in the method design: our approach induces different behaviors by optimizing opposite optimization objectives on the base set and the poisoned set, whereas confusion training relies on random labeling to disrupt the learning of the clean samples. Importantly, we design a nested procedure that can effectively deal with the failure cases of \cite{qi2022fight}. Moreover, our method distinguishes itself from \cite{qi2022fight} by providing additional important advantages: (1) our approach does not require the poisoned set to be labeled, thereby enabling applications in SSL settings; and (2) our approach is robust to different poison ratios without ratio-specific tuning and can effectively detect attacks generating triggers entangled with semantic features.



\vspace{-1.em}
\section{Attacker \& Defender Models}
\vspace{-.5em}
\label{sec:threatmodel}
This section discusses standard threat models and assumptions about defender knowledge for different DL paradigms.

\noindent \textbf{\underline{Case-0} End-to-end SL}:
In this setting, the attacker performs the backdoor attack by injecting a set of poisoned samples into the training dataset. The defender has access to the poisoned training dataset and the downstream learning algorithm. The defender's goal is to identify the poisoned samples within the training set and further remove the identified samples to prevent backdoor attacks from taking effect.


\noindent \textbf{\underline{Case-1} SSL Adaptation}: 
Under this setting, the attacker performs the backdoor attack by poisoning the unlabeled dataset \cite{saha2022backdoor, li2022demystifying}. Following prior attack literature, we assume that the attacker does not have
access to the fine-tuning task---the dataset or algorithm. Thus, the dataset used for fine-tuning is clean, and the attack only affects the unlabeled dataset. The defender has access to the complete training data, including both the data for SSL as well as the data for fine-tuning. In addition, the defender knows the algorithm for SSL and fine-tuning. The goal of the defender is to identify and remove the poisoned samples from the unlabeled dataset. 
Other attack settings target multi-modal contrastive learning, such as attacking the CLIP \cite{carlini2021poisoning}, is not considered in this case, as training CLIP requires additional text input supervision \cite{radford2021learning}.


\noindent \textbf{\underline{Case-2} Transfer Learning}: The attacker performs the backdoor attack by poisoning a labeled dataset used for fine-tuning an existing pre-trained model. The defender knows the pre-trained model, the entire fine-tuning dataset (whose size often cannot support training a model from scratch), as well as the fine-tuning algorithm. The goal of the defender is to detect the poisoned samples within the fine-tuning dataset.

In all three cases, we assume the attacker can poison no more than half of the training dataset. We also assume that the defender has a small set of clean, unlabeled samples (the base set) to help with detection. These clean samples can be manually or automatically screened \cite{zeng2022sift}. Compared with most recent detection methods \cite{qi2022fight}, which require a \emph{labeled} clean base set of at least 2000 samples, our method \emph{relaxes} the requirement on the label information.
\vspace{-1.em}
\section{Proposed Method}


\vspace{-1.em}
\subsection{Key Idea}
\vspace{-.5em}
\label{sec:keyidea}

\setlength{\intextsep}{1em}%
\setlength{\columnsep}{1em}%
\begin{wrapfigure}{r}{0.6\columnwidth}
\vspace{-3em}
  \begin{center}
    \includegraphics[width=.55\columnwidth]{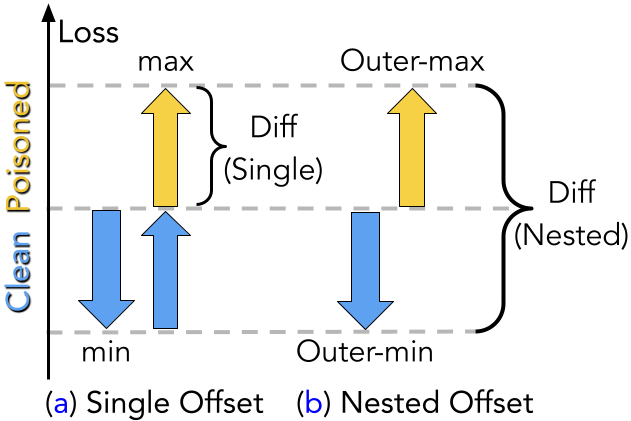}
  \end{center}
 \vspace{-1.em}
  \caption{Illustration of \textbf{(\textcolor{blue}{a})} single offset procedure and \textbf{(\textcolor{blue}{b})} how the power of differentiating between clean and poison improves when the single offset is replaced by a nested loop. }
   \label{fig:nested}
  \vspace{-1em}
\end{wrapfigure}
Our goal is to enforce distinguishable model behaviors on poisoned and clean samples actively. 
The key idea is to design two optimizations that induce \emph{opposite} model behaviors on \emph{the poisoned dataset} (including its clean and poisoned portion) and \emph{the clean base set}. Specifically, the two optimizations are performed simultaneously, where the first one \emph{minimizes} a certain loss function on the clean base set and the second one \emph{maximizes} the same loss on the entire poisoned training dataset. Note that the clean portion of the poisoned dataset and the clean base set are both drawn from the same clean distribution. Hence, the effect of the second optimization on the clean samples will be \emph{offset} by the first optimization, and the loss on clean samples after the two optimizations is closer to the loss before. 
By contrast, the poisoned samples only go through the second optimization; therefore, the loss on the poisoned samples is maximized. Overall, as a result of the two optimizations, poisoned and clean samples will produce different loss values, thus becoming separable. The single offset's effect on clean samples and poisoned samples is illustrated in Figure~\ref{fig:nested} (\textbf{\textcolor{blue}{a}}).

\noindent
\textbf{Intuition on the distinguishability of poisons.} 
Poisoning, whether through additive triggers~\cite{gu2017badnets}, generative models~\cite{li2021invisible}, affine transformations~\cite{nguyen2021wanet}, or even adaptive perturbation techniques~\cite{zeng2021rethinking}, introduces a distributional shift from clean data. The resulting poisoning distribution and the original clean distribution have disjoint support, and thus the total variation (TV) distance between the two distributions is one. The Le Cam's lower bound, a classic result in statistical learning (refer to Chapter 15 in~\cite{wainwright_2019}), states that the minimum error over all detectors that classify the samples from two distributions, $P_1$ and $P_2$, is equal to $1/2(1- \|P_1-P_2\|_\text{TV})$. Hence, there exists a detector achieving zero error probability for distinguishing between poisons and non-poisons. Le Cam's bound guarantees the existence of a good detector as long as poisons do not naturally appear in the clean distribution, and our method to be introduced is an effort to find such a detector based on the information of a clean base set.

\vspace{-1.em}
\subsection{Detection via Offset}
\vspace{-.5em}
Now, we formalize the offset idea for poisoned sample detection.
Let $D_\text{b}$ denote the clean base set and $D_\text{poi}$ denote the poisoned training set. Formally, we can characterize the process of inducing distinguishable behaviors on poisoned and clean samples as a multi-objective optimization:
\vspace{-.8em}
\begin{align}
\theta^* \in \arg\min_{\theta}&\frac{1}{\left | D_\text{b} \right |}\sum_{x_\text{b} \in D_\text{b}} \mathcal{L}_\text{min} \left(f(x_\text{b}|\theta)\right)\nonumber\\
&-\frac{1}{\left | D_\text{poi} \right |}\sum_{x_\text{poi} \in D_\text{poi}} \mathcal{L}_\text{max} \left(f(x_\text{poi}|\theta)\right).
\label{eqn:offset}
\end{align}

\vspace{-.8em}
\noindent
When discussing the high-level idea of our method, we assume that the minimization and maximization employ the same objective, i.e., $\mathcal{L}_\text{min}=\mathcal{L}_\text{max}$. However, these two functions can also be different; as long as minimizing $\mathcal{L}_\text{min}$ and maximizing $\mathcal{L}_\text{max}$ induce different model behaviors, one optimization will mitigate the effect of the other on the clean distribution. 

In the implementation, we do not directly solve the optimization with two optimizations at the same time due to the instability of the corresponding optimization path; instead, we \emph{loop} between two objectives:
\begin{enumerate}
\vspace{-.7em}
    \item We first minimize $\mathcal{L}_\text{min}$ by taking a \emph{gradient descent} step on a mini-batch drawn from the base set;
    \vspace{-1.em}
    \item Then, we utilize the resulting model as the initializer for maximizing $\mathcal{L}_\text{max}$ and perform a \emph{gradient ascent} step on a mini-batch drawn from the poisoned set;
    \vspace{-1.em}
    \item Repeat the above two steps.
    \vspace{-.7em}
\end{enumerate}
We empirically observe the alternating procedure is stable. As the focus of the paper is to develop practical detection methods, we will defer the theoretical analysis of this procedure---an interesting open problem---for future work.
    

Next, we will discuss which loss function we shall use to instantiate $\mathcal{L}_\text{min}$ and $\mathcal{L}_\text{max}$. With the goal of detecting unlabeled poisoned data in mind, we propose a loss function, which calculates the variance of the logits. Let $f(x|\theta)$ denote the output logit of a model that is parameterized by $\theta$ and takes $x$ as input. For a model that performs $k$-class classification, $f(x|\theta)\in \mathbb{R}^k$ and the $i$-th class logit is denoted by $f(x|\theta)_i$. Furthermore, let $\overline{f(x|\theta)}$ denote the average of the output logits of all classes. Then, our proposed loss function can be expressed as
\vspace{-.8em}
\begin{equation}
\mathcal{L}_\text{var}(f(x|\theta)) = \frac{1}{k}\sum_{i=0}^{k} \bigg (f(x|\theta)_i - \overline{f(x|\theta)} \bigg )^2.
\vspace{-.8em}
\end{equation}
When the detection is performed on the \emph{unlabeled} data, we can instantiate both $\mathcal{L}_\text{min}$ and $\mathcal{L}_\text{max}$ to be $\mathcal{L}_\text{var}$ defined above, because calculating $\mathcal{L}_\text{var}$ does not require label information. As the result of minimizing $\mathcal{L}_\text{min}$, the clean samples are forced to have a \emph{flat} logit pattern.
Then, the maximization optimization maximizes the same loss on the poisoned dataset, which induces high-variance logits for poisoned samples. For clean samples, the effects of maximization and minimization are roughly canceled out.
Therefore, clean samples are expected to produce lower-variance logits than poisoned samples.

When the detection is performed on a labeled poisoned dataset,
we find that instantiating $\mathcal{L}_\text{max}$ with the cross-entropy-based prediction loss $\mathcal{L}_\text{ce}$ achieves a good detection performance faster than $\mathcal{L}_\text{var}$: 
\vspace{-1.em}
\begin{align}
    \mathcal{L}_\text{ce}(f(x|\theta),y) = -\sum_{i=1}^k y_i \log \sigma(f(x|\theta))_i,
\end{align}

\vspace{-1.em}
\noindent
where $\sigma(x)_i$ denotes the $i$-th output of the softmax and $y$ represents the one-hot encoding of $x$'s label. 

It is worth mentioning that we fix the minimization loss to be $\mathcal{L}_{\text{var}}$ regardless of whether the base set is labeled or unlabeled. We found that even when the label information is available, this choice still leads to better detection performance than using $\mathcal{L}_{\text{ce}}$ as the minimization goal. This is because learning through minimizing $\mathcal{L}_{\text{var}}$ will make the model extract class-independent features. 
A mini-batch of the base set may be class-imbalanced or sometimes contain only partial classes due to random sampling.
Hence, $\mathcal{L}_{\text{var}}$ can be more steadily minimized than $\mathcal{L}_{\text{ce}}$ via mini-batch gradients.

\vspace{-1.3em}
\subsection{
Strengthened Detection via Nested Offset
}
\vspace{-.7em}
\label{sec:discussouter}


\noindent
\textbf{Weakness of a single offset.} Despite the neatness of the offset idea, directly solving the two optimizations with the proposed loss functions is limited in tackling attacks with \emph{low poison ratio} and the settings where \emph{poisoned samples take effect slowly during training} (i.e., attacks need many epochs of training to obtain a high enough success rate; examples of such attacks include \cite{zeng2022narcissus,li2022demystifying}). The reasons are as follows. In the low poison ratio setting, mini-batches naturally contain very small amounts of poisoned samples; on the other hand, each gradient ascent step takes a step towards reducing the \emph{average} loss over a mini-batch and tends to overlook the minorities. Hence, the loss of poisoned samples would be increased by less with a lower poison ratio. To explain the second limitation, note that $\theta$ is an over-parameterized model (e.g., ResNet-18 and Vision Transformer). If an attack takes many epochs to take effect, then we need to train $\theta$ for long enough. The model after long training will end up ``memorizing'' all the samples from the base set and the poisoned set, i.e., all the samples from the base set achieve a low value of $\mathcal{L}_{\text{min}}$ and all the samples from the poisoned set (including both clean and poisoned samples) to achieve a high value of $\mathcal{L}_{\text{max}}$. In that case, the poisoned portion and the clean one are inseparable.

\noindent
\textbf{How to mitigate these failure cases?} To illustrate our idea, let us think about a hypothetical design, assuming one can perfectly pinpoint a set of poisoned samples. In this design, we keep the first step minimizing on the clean base set, but the second maximization is performed on purely poisoned samples instead of the poisoned training set, which generally contains a large portion of clean samples and only a small portion of poisoned samples. This hypothetical design would be able to solve the two failure cases above. For the first case, since mini-batches for maximization contain solely poisoned samples, the poisoned samples would still have their loss increased and thus is distinguishable from the clean ones. For the second case, while long training can lead to memorization but with the hypothetical design, it is just the poisoned samples that get memorized and are assigned with high loss; therefore, the poisoned samples and the clean ones are still separable.

While having access to a set of purely poisoned samples is not realistic, this thought experiment inspires an idea to improve an offset-based detection approach, which is to replace the poisoned training set (dominated by clean samples) with a set dominated by poisoned samples in the second maximization. To form such a poison-dominated set, we can leverage a new offset loop (referred to as the \emph{inner} offset loop) to mark a set of the most suspicious samples. Then, we use those samples to perform maximization 
of the original offset loop (referred to as the \emph{outer} offset loop).


\noindent
\textbf{How to design the inner offset loop that provides a poison-condensed set?} First, it is not ideal to reuse the design of the outer loop for this inner one, because in that case the inner would suffer the same ``memorization'' issue. Instead, we aim to avoid ``overparameterized'' models and perform the inner loop with a simple model. On the other hand, a simple model could be incapable of extracting complex features to support the detection of poisoned samples. Our solution is to use the poisoned model (i.e., the downstream model trained on the poisoned dataset) to extract features from the poisoned set and the base set and then optimize a simple model to detect the poisoned samples in the feature space.

Note that the embedding space of a poisoned model has been shown to be informative to detect many but not all backdoor attacks (detailed in Section \ref{sec:evaluation}).
Although the poisoned and clean samples are not perfectly separable based on the embeddings---as 
 illustrated in Figure \ref{fig:tsne}---the reason why these methods underperform in many cases, the poisoned model still provides a well-trained embedding space and some imperfect signals for selecting a poison-condensed set.

\setlength{\columnsep}{1em}%

\noindent\textbf{Detailed design of the inner offset loop.} The inner offset loop is executed \emph{inside} the previous offset loop (Eqn.~\ref{eqn:offset}). It condenses the poison in a mini-batch sampled by the maximization step of the outer offset loop.
Specifically, the inner offset loop will return a set of samples marked as poison. 
We will use this poison-condensed subset of the original mini-batch to perform the outer maximization. When the inner loop is relatively precise in gathering a poison-condensed subset, the outer loop will maximize the outer loss of poisoned samples without introducing much offset effect on clean samples. As a result, the poisoned and clean samples become more distinguishable in terms of the outer loss compared to a single offset loop via Eqn.~\ref{eqn:offset}. An intuitive explanation of the improvement is illustrated by Figure~\ref{fig:nested} (\textbf{\textcolor{blue}{b}}).

Let $f(x|\theta_\text{poi}^*)$ denote the poisoned model, and its parameters are given by $\theta_\text{poi}^*$. Let $M(\cdot|w)$ be a mapping from the logits to a real value in the range $[0,1]$, and $w$ denotes its parameters. The inner offset can be characterized by
\vspace{-1.em}
\begin{align}
    w^* &= \arg\min_{w}\underbrace{\frac{1}{\left | B_\text{b} \right |}\sum_{x_\text{b} \in B_\text{b}} \mathcal{L}_\text{BCE} \left(M(f(x_\text{b}|\theta_\text{poi}^*)|w),0\right)}_{\mathcal{L}_1}\nonumber\\
&+\underbrace{\frac{1}{\left | B_\text{poi} \right |}\sum_{x_\text{poi} \in B_\text{poi}} \mathcal{L}_\text{BCE} \left(M(f(x_\text{poi}|\theta_\text{poi}^*)|w),1\right)}_{\mathcal{L}_2},
\end{align}

\vspace{-1.em}
\noindent
where $\mathcal{L}_\text{BCE}(p,q) = -p\log q + (1-p)\log (1-q)$, representing the binary cross entropy loss and $B_\text{b}$ and $B_\text{poi}$ stand for a mini-batch drawn from the clean base set and the poisoned training set, respectively. 

The first minimization objective will encourage learning a mapping $M$ such that the mini-batch from the clean base set is labeled as ``$0$''; the second objective will further promote $M$ to label the mini-batch from the poisoned set as ``$1$''. By minimizing the two objectives simultaneously, the effect on the clean data gets canceled. As a result, the clean samples will be predicted as ``$1$'' with low confidence, yet the poisoned ones will be predicted as ``$1$'' with high confidence. Then, we can mark the samples with the highest confidence or the lowest BCE loss for predicting ``$1$'' as the suspicious poisoned samples. In practice, $M$ is implemented as a two-layer, full-connected network with 128 hidden neurons.
Again, to avoid stability issues, in the implementation, we first take a gradient descent step to minimize $\mathcal{L}_1$ and then take a gradient ascent step to minimize $\mathcal{L}_2$, and alternate between the two steps. 

The pseudo-code for the inner offset loop is provided in Algorithm~\ref{algo:concentration}, termed \emph{Poison Concentration}.
\vspace{-1.em}

\SetKwInput{KwParam}{Parameters}
\begin{algorithm}[!h]
\algsetup{linenosize=\tiny}
\small
    \caption{Poison Concentration}
    \label{algo:concentration}
    \SetNoFillComment
    \KwIn{
    $\theta_{\text{poi}}^*$ (Poisoned feature extractor);
    \\ \quad \quad \quad 
    $B_{\text{poi}}$ (Poisoned training mini-batch);
    \\ \quad \quad \quad 
    $B_\text{b}$ (Base set mini-batch);
    }
    \KwOut{$B_{\text{pc}}$ (Poison concentrated mini-batch);}
    \KwParam{
    $\mathcal{N}$ (Total inner loop iteration number);
    \\  \quad \quad \quad  \quad \quad \quad 
    $\gamma > 0$ (Step size);
    \\  \quad \quad \quad  \quad \quad \quad 
    $\lambda$ (Threshold);
    }
    \BlankLine
    \tcc{1.Dynamic training of $M$}
    \For{each iteration $j$ in $(0,\mathcal{N}-1)$}{

        $M^{\prime}_j \leftarrow 
        M_{j} - \gamma
        \frac{1}{|{B}_\text{b}|}
        \sum_{x_\text{b}\in{B}_\text{b}}
        \frac{\partial\mathcal{L}_{\text{BCE}}\left(M\left(f(x_\text{b}|\theta_{\text{poi}}^*)\right),0
        \right)
        }{\partial M
        }$
        \;
        
        $M_{j+1} \leftarrow 
        M_{j}^{\prime} - \gamma
        \frac{1}{|{B}_{\text{poi}}|}
        \sum_{x_{\text{poi}}\in{B}_{\text{poi}}}
        \frac{\partial\mathcal{L}_{\text{BCE}}\left(M^{\prime}\left(f(x_{\text{poi}}|\theta_{\text{poi}}^*)\right),1\right)
        }{\partial M^{\prime}
        }$\;
      }
      
      \tcc{2.Get output values}
      $V \leftarrow M_{\mathcal{N}}\left(f(B_{\text{poi}}|\theta_{\text{poi}}^*)\right)$\;
      \tcc{3.Using AO to determine outliers}
      $B_{\text{pc}} \leftarrow B_{\text{poi}} [\text{\textbf{AO}}\left(V\right)\geq \lambda]$\;
      \Return $B_{\text{pc}}$
\end{algorithm}

\vspace{-1.em}
\noindent
\textbf{Adaptive thresholding for the inner offset.} The last step of Poison Concentration is to select the subset marked as poison based on the confidence score output by $M$. We will elaborate on how to adaptively choose the size of this subset. First, directly adopting a fixed threshold to identify the most likely poisoned samples is impractical, as different mini-batches may contain different amounts of poisons.
To tackle this problem, we adopt Adjusted Outlyingness (AO) \cite{brys2005robustification} to adaptively determine the number of most suspicious samples within each mini-batch. AO maps the BCE losses into a scale such that a fixed threshold can effectively identify the most suspicious samples. 
Note that AO does not aim to filter out as many poisoned samples as possible within the mini-batch; instead, it is adopted to achieve high precision, i.e., identifying a subset of the mini-batch that is dominated by poisoned samples. 
In the evaluation, we threshold the output of AO with 2. By the nature of AO, we are essentially adopting an adaptive threshold despite using a fixed output value (see Figure \ref{fig:mentoutput}).

\vspace{-1.em}
\subsection{Overall Workflow 
}
\vspace{-.5em}
The overall algorithm of $\AlgName$ with two offset loops is presented in Algorithm \ref{algo:mainAlgo}. Functionally speaking, the inner loop condenses the poison within each mini-batch drawn from the poisoned dataset, the outer loop induces different model behaviors on clean samples and poisoned samples.
At each iteration of the outer, we minimize $\mathcal{L}_\text{var}$ by taking mini-batch gradient descent with samples from the clean base set; then, we perform the poison concentration step: the inner returns subset of samples most likely to be poisoned; then proceed to the maximization step of the outer $\mathcal{L}_\text{max}$ by doing gradient ascent with the suspicious points returned by the inner. In the end, we can obtain a detector model $f(\cdot|\theta_{\mathcal{I}})$ with parameters $\theta_{\mathcal{I}}$ obtained after $\mathcal{I}$ outer iterations and this model induces different values of $\mathcal{L}_{\text{max}}$ between clean and poisoned samples.

\vspace{-1.em}
\SetKwInput{KwParam}{Parameters}
\begin{algorithm}[!h]
\algsetup{linenosize=\tiny}
\small
    \caption{$\AlgName$ Backdoor Detection}
    \label{algo:mainAlgo}
    \SetNoFillComment
    \KwIn{
    $\theta_0$ (Initialized detector);
    \\ \quad \quad \quad \,
    $\theta_{\text{poi}}^*$ (Poisoned feature extractor);
    \\ \quad \quad \quad 
    $D_{\text{poi}}$ (Poisoned training set);
    \\ \quad \quad \quad 
    $D_\text{b}$ (Base set);
    }
    \KwOut{$S_{\text{poi}}$ (Indexes of the detected poisoned samples);}
    \KwParam{
    $\mathcal{I}$ (Total outer loop iteration number);
    \\  \quad \quad \quad  \quad \quad \quad 
    $\alpha > 0$ (Step size);
    }
    \BlankLine
    \For{each iteration $i$ in $(0,\mathcal{I}-1)$}{
          \tcc{1. Obtaining mini-batches}
          $B^i_{\text{poi}} \leftarrow B^i_{\text{poi}} \in D_{\text{poi}}$\;
          $B^i_\text{b} \leftarrow B^i_\text{b} \in D_\text{b}$\;
          
          \tcc{2. Minimization}
          $\theta^{\prime}= \leftarrow 
            \theta_{i} - \alpha
            \frac{1}{|{B}^i_\text{b}|}
        \sum_{x^i_\text{b}\in{B}^i_\text{b}}
        \frac{
            \partial\mathcal{L}_{\text{var}}\left(f(x^i_\text{b}|\theta_i)\right)}{\partial \theta_i
        }
            $\;
            
        \tcc{3. Poison Concentration}
        $B^i_{\text{pc}} \leftarrow \text{\textbf{Poison Concentration}}\left(B^i_{\text{poi}}, B^i_{\text{b}}, \theta_{\text{poi}}^*\right)$ \;
        
          \tcc{4. Maximization}
          $\theta_{i+1} \leftarrow 
            \theta^{\prime}_{i} + \alpha
            \frac{1}{|{B}^i_{\text{pc}}|}
        \sum_{x^i_{\text{pc}}\in{B}^i_{\text{pc}}}
        \frac{
            \partial
            \mathcal{L}_{\text{max}}\left(f(x^i_{\text{pc}}|\theta^{\prime})\right)
            }{\partial \theta^{\prime}
        }
            $\;
      }
      \tcc{5.Get output loss values}
      $V \leftarrow \mathcal{L}_{\text{max}}\left(f(D_{\text{poi}}|\theta_{\mathcal{I}})\right)$\;
      \tcc{6.Detection result via adaptive GMM}
      $S_{\text{poi}} \leftarrow\text{\textbf{adaptive GMM}}\left(V\right)$\;
      \Return $S_{\text{poi}}$
\end{algorithm}

\vspace{-1.em}


\noindent
\textbf{Adaptive thresholding for the outer loop.} With the trained detector model, $\theta_{\mathcal{I}}$, we now discuss how to identify the poisoned samples. Similar to the inner, we propose an adaptive thresholding method for the outer as well. Note that the threshold of the inner and outer loop has distinct goals. The inner loop aims to identify a subset with a high density of poisons, while the outer loop aims to adaptively conduct a split between the clean and poisoned loss distribution that helps the detector to remove as many poisons as possible while maintaining a low false positive, i.e., high precision is prioritized for the inner yet high recall is prioritized for the latter.

\begin{figure}[t!]
  \centering
     \vspace{-2em}
  \includegraphics[width=0.85\linewidth]{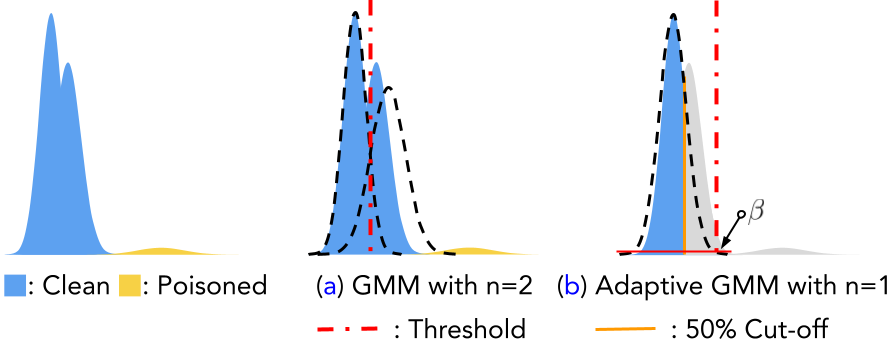}
  \vspace{-1.em}
  \caption{
Illustration of \textbf{(\textcolor{blue}{a})} the problem of GMM over long-tailed cases where the attacks are of low poison ratio and \textbf{(\textcolor{blue}{b})} how the proposed adaptive GMM can help.
  }
  \label{fig:gmm}
  \vspace{-1.8em}
\end{figure}


As will be shown later, after the overall optimization, $f(\cdot|\theta_{\mathcal{I}})$
will output distinct loss distribution for the clean and poisoned samples. One might be tempted to directly fit a Gaussian Mixture Model (GMM) with two components.
However, doing so is problematic, as depicted in Figure \ref{fig:gmm}. 
Since there are usually much fewer poisoned samples than clean ones, the GMM tends to split the multiple-modal clean distribution into two Gaussian distributions instead of fitting two Gaussians respectively to the clean and poison distributions.


To tackle this problem, we propose a simple twist of GMM, termed \textbf{adaptive GMM}. We first abandon half of the samples achieving the highest values of $\mathcal{L}_{\text{max}}$, which will remove all the poisoned samples (we assume the attacker can poison no more than half of the training dataset, Section \ref{sec:threatmodel}). Then, we fit a Gaussian to the remaining points. Since the optimized detector model largely centers the clean samples' loss close to $\mathcal{L}_{\text{var}}=0$ or $\mathcal{L}_{\text{ce}}=-\log(\frac{1}{k})$, the Gaussian fitted on the remaining samples remains similar to the Gaussian fitted on all the non-poisons
(see Figure \ref{fig:gmm} (b)). 
Lastly, we set a small threshold on the Gaussian density, $\beta$, to cut off the samples that are unlikely to be generated from the fitted Gaussian. In practice, we set the cut-off threshold as $\beta=10^{-6}$, which equivalently keeps the lowest-loss samples with a probability higher than $>99.99\%$ being generated from the fitted Gaussian (for any Gaussian distribution with a variance smaller than $10$).

\vspace{-1.5em}
\section{Evaluation}
\vspace{-1em}
\label{sec:evaluation}
%
Our evaluation aims to answer the following questions.
\begin{compactitem}
    \item \textbf{\underline{Case-0}} (Section \ref{sec:case0exp}): How does $\AlgName$ compare with other methods in end-to-end SL setting? Is detection effective when multiple attacks exist simultaneously? How does the detection performance vary over different attacks and poison ratios?  
    \item \textbf{\underline{Case-1}} (Section \ref{sec:sslexp}): Can $\AlgName$ robustly detect attacks in SSL settings? 
    How does the knowledge about downstream tasks affect the defense's effect?
    \item \textbf{\underline{Case-2}} (Section \ref{sec:finetuneexp}): Can $\AlgName$ provide reliable backdoor sample detection in TL settings? 
    What are the limitations of other defenses in this setting?
    \item \textbf{Adaptive Attack} (Section \ref{sec:adptiveattack}): Is it possible to adaptively evade $\AlgName$'s detection?
    \item \textbf{Ablation Study} (Appendix \ref{sec:ablation}): How do different design choices affect the final performance of $\AlgName$?
\end{compactitem}

\vspace{-1.2em}
\subsection{Settings}
\label{sec:settings}
\vspace{-0.8em}

\noindent
\textbf{Evaluation metrics.} There are two key aspects throughout our evaluation: (1) How accurately can the poisoned samples be detected (\textbf{\emph{upstream evaluation}})? (2) After the suspicious points are removed, how would a downstream model learn from the remaining data perform (\textbf{\emph{downstream evaluation}})?

For upstream evaluation, we utilize two metrics, namely, True Positive Rate (\textbf{TPR}), $TPR = {TP}/{(TP+FN)}$,
and False Positive Rate (\textbf{FPR}), $FPR = {FP}/{(FP+TN)}$,
where $TP$, $FP$, $TN$, and $FN$ denote the number of true positives, false positives, true negatives, and false negatives, respectively\footnote{Note that poison is considered positive and clean is considered negative.}.
TPR depicts how well a specific backdoor detection method filters out the backdoored samples. A higher TPR (closer to 100$\%$) denotes a stronger filtering ability. 
FPR depicts how precise the filtering is: when a specific method achieves TPR that is high enough, FPR helps us to understand the trade-off, i.e., how many clean samples are wasted and wrongly flagged as backdoored during the detection. A lower FPR shows that fewer clean samples are wasted, and more clean data shall be kept and available for downstream usage.

One thing worth noting is that no detection method can reliably remove all the poisoned samples. However, \emph{the remained backdoor samples that go unnoticed by a successful defense should be small enough to deactivate attacks}.
Thus, we evaluate the backdoor attacks' Attack Success Rate (\textbf{ASR}) on the downstream model trained using the filtered dataset to study whether the detection is good enough to stop attacks. ASR measures the proportion of backdoored test samples being classified into target classes. Additionally, we evaluate the downstream model's Clean Accuracy (\textbf{ACC}). A high ACC means that the detection method is able to maintain a large enough clean set to support the model performance. 


\begin{table*}[t!]
    \vspace{-1.5em}
    \centering
    \resizebox{0.9\textwidth}{!}{
    \begin{tabular}{cp{0.96cm}<{\centering}p{0.96cm}<{\centering}|p{0.96cm}<{\centering}p{0.96cm}<{\centering}|p{0.96cm}<{\centering}p{0.96cm}<{\centering}|p{0.96cm}<{\centering}p{0.96cm}<{\centering}|p{0.96cm}<{\centering}p{0.96cm}<{\centering}|p{0.96cm}<{\centering}p{0.96cm}<{\centering}|p{0.96cm}<{\centering}p{0.96cm}<{\centering}
    ||cc||cc}
    \hline
    \cellcolor[HTML]{FFFFFF}{\color[HTML]{000000} } &
      \multicolumn{8}{|c|}{\cellcolor[HTML]{FFFFFF}{\color[HTML]{000000} \textit{Dirty-Label Backdoor Attacks}}} &
      \multicolumn{6}{c||}{\cellcolor[HTML]{FFFFFF}{\color[HTML]{000000} \textit{Clean-Label Backdoor Attacks}}} 
      & \multicolumn{2}{c||}{\multirow{2}{*}{\textbf{Average}}}
      & \multicolumn{2}{c}{\multirow{2}{*}{\textbf{Worst-Case}}}
      \\ \cline{2-15} 
    \cellcolor[HTML]{FFFFFF}{\color[HTML]{000000} } &
      \multicolumn{2}{|c|}{\textbf{BadNets (5\%)}} &
      \multicolumn{2}{c|}{\textbf{Blended (5\%)}} &
      \multicolumn{2}{c|}{\textbf{WaNet (10\%)}} &
      \multicolumn{2}{c|}{\textbf{ISSBA (1\%)}} &
      \multicolumn{2}{c|}{\textbf{LC (1\%)}} &
      \multicolumn{2}{c|}{\textbf{SAA (1\%)}} &
      \multicolumn{2}{c||}{\textbf{Narci. (0.05\%)}} & & \\ 
\hline
\multicolumn{19}{c}{
    \multirow{2}{*}{\textbf{(\textcolor{blue}{a}) Upstream Evaluation}}} \\ \\
       \hline

    \multicolumn{1}{c|}{}&
       \multicolumn{1}{c}{TPR} $\uparrow$&
      FPR $\downarrow$&
      TPR $\uparrow$&
      FPR $\downarrow$&
      TPR $\uparrow$&
      FPR $\downarrow$&
      TPR $\uparrow$&
      FPR $\downarrow$&
      TPR $\uparrow$&
      FPR $\downarrow$&
      TPR $\uparrow$&
      FPR $\downarrow$&
      TPR $\uparrow$&
      FPR $\downarrow$&
      TPR $\uparrow$&
      FPR $\downarrow$&
      TPR $\uparrow$&
      \multicolumn{1}{|c}{FPR $\downarrow$}\\ \hline

    \multicolumn{1}{c|}{\textbf{Spectral}} &
      95.6 &
      2.86 &
      99.8 &
      2.64 &
      0.64 &
      \textcolor{minired}{16.6} &
      \textcolor{minired}{0.00} &
      1.52 &
      80.2 &
      0.71 &
      87.6 &
      0.63 &
      \textcolor{minired}{0.00} &
      0.08 &
      51.9 &
      3.58 & 
      0.00 &
      \multicolumn{1}{|c}{16.6} \\ \hline
      
    \multicolumn{1}{c|}{\textbf{Spectre}} &
      96.9 &
      0.28 &
      99.8 &
      2.64 &
      1.00 &
      \textcolor{minired}{16.6} &
      80.2 &
      0.71 &
      99.8 &
      0.51 &
      99.4 &
      0.51 &
      \textcolor{minired}{0.00} &
      0.07 &
      68.2 &
      3.05 &
      0.00 &
      \multicolumn{1}{|c}{16.6} \\ \hline
      
    \multicolumn{1}{c|}{\textbf{Beatrix}} &
      93.8 &
      1.81 &
      67.9 &
      \textcolor{minired}{3.04} &
      82.2 &
      0.53 &
      73.4 &
      1.31 &
      91.2 &
      0.29 &
      69.8 &
      1.58 &
      \textcolor{minired}{12.0} &
      1.97 &
      70.4 &
      1.50 &
      12.0 &
      \multicolumn{1}{|c}{3.04} \\ \hline
      
    \multicolumn{1}{c|}{\textbf{AC}} &
      90.5 &
      40.1 &
      65.4 &
      \textcolor{minired}{44.9} &
      8.30 &
      41.5 &
      11.0 &
      41.1 &
      91.2 &
      0.41 &
      75.6 &
      21.5 &
      \textcolor{minired}{0.00} &
      34.3 &
      48.9 &
      32.0 &
      0.00 &
      \multicolumn{1}{|c}{44.9} \\ \hline

    \multicolumn{1}{c|}{\textbf{ABL}} &
      85.4 &
      3.40 &
      93.4 &
      2.98 &
      28.1 &
      \textcolor{minired}{13.5} &
      55.2 &
      0.96 &
      87.2 &
      0.63 &
      73.4 &
      0.77 &
      \textcolor{minired}{0.00} &
      0.07 &
      60.4 &
      3.19 &
      0.00 &
      \multicolumn{1}{|c}{13.5} \\ \hline

    \multicolumn{1}{c|}{\textbf{Strip}} &
      25.4 &
      11.5 &
      17.3 &
      12.1 &
      5.08 &
      \textcolor{minired}{10.0} &
      68.8 &
      9.34 &
      100&
      0.85 &
      63.4 &
      1.22 &
      \textcolor{minired}{0.00} &
      0.05
      & 40.0 &
      6.44 &
      0.00 &
      \multicolumn{1}{|c}{10.0} \\ \hline

    \multicolumn{1}{c|}{\textbf{CT}} &
      99.0 &
      3.72 &
      98.1 &
      4.53 &
      95.8 &
      2.64 &
      96.6 &
      4.37 &
      100 &
      \textcolor{minired}{9.01} &
      95.2 &
      5.44 &
      \textcolor{minired}{0.00} &
      5.54 
      &83.5 &
      5.03 &
      0.00 &
      \multicolumn{1}{|c}{9.01}\\ \hline

    \multicolumn{1}{c|}{\textbf{Ours}} &
      99.5 &
      0.55 &
      100 &
      0.00 &
      \textcolor{minired}{90.7} &
      \textcolor{minired}{8.09} &
      95.6 &
      0.36 &
      96.2 &
      0.75 &
      96.6 &
      0.39 &
      92.0 &
      0.34 &
      \textbf{95.8} &
      \textbf{1.49} &
      \textbf{90.7} &
      \multicolumn{1}{|c}{\textbf{8.09}} \\ \hline
      
    \multicolumn{19}{c}{
\multirow{2}{*}{\textbf{(\textcolor{blue}{b}) Downstream Evaluation}}} \\ \\
   \hline    

      \multicolumn{1}{c|}{} &
        ASR $\downarrow$ &
        ACC $\uparrow$ &
        ASR $\downarrow$ &
        ACC $\uparrow$ &
        ASR $\downarrow$ &
        ACC $\uparrow$ &
        ASR $\downarrow$ &
        ACC $\uparrow$ &
        ASR $\downarrow$ &
        ACC $\uparrow$ &
        ASR $\downarrow$ &
        ACC $\uparrow$ &
        ASR $\downarrow$ &
        ACC $\uparrow$ &
        ASR $\downarrow$ &
        ACC $\uparrow$&
        ASR $\downarrow$ &
        \multicolumn{1}{|c}{ACC $\uparrow$} \\ \hline

     \multicolumn{1}{c|}{\textbf{No Def.}} &
        \cellcolor[HTML]{F3CECA}96.5 &
        \multicolumn{1}{c|}{\textcolor{minired}{93.4}} &
        \cellcolor[HTML]{F3CECA}94.9 &
        \multicolumn{1}{c|}{93.5} &
        \cellcolor[HTML]{F3CECA}99.4 &
        \multicolumn{1}{c|}{93.5} &
        \cellcolor[HTML]{F3CECA}92.6 &
        94.1 &
        \cellcolor[HTML]{F3CECA}\textcolor{minired}{100} &
        \multicolumn{1}{c|}{94.7} &
        \cellcolor[HTML]{F3CECA}76.7 &
        \multicolumn{1}{c|}{{94.4}} &
        \cellcolor[HTML]{F3CECA}99.7 &
        {94.9} 
        &94.3 & 94.1 & 100 & \multicolumn{1}{|c}{93.4}
        \\ \hline
        \hline

    \multicolumn{1}{c|}{\textbf{Spectral}} &
        \cellcolor[HTML]{F3CECA}48.4 &
        \multicolumn{1}{c|}{94.5} &
        \cellcolor[HTML]{D5E2F1}10.7 &
        \multicolumn{1}{c|}{94.1} &
        \cellcolor[HTML]{F3CECA}98.9 &
        \multicolumn{1}{c|}{90.0} &
        \cellcolor[HTML]{F3CECA}93.0 &
        94.1 &
        \cellcolor[HTML]{D5E2F1}10.6 &
        \multicolumn{1}{c|}{\textcolor{minired}{94.8}} &
        \cellcolor[HTML]{D5E2F1}3.11 &
        \multicolumn{1}{c|}{94.2} &
        \cellcolor[HTML]{F3CECA}\textcolor{minired}{99.7} &
        \textcolor{minired}{94.8} 
        &52.1 & 93.8 & 99.7 & \multicolumn{1}{|c}{90.0} \\ \hline
        
    \multicolumn{1}{c|}{\textbf{Spectre}} &
        \cellcolor[HTML]{F3CECA}34.8 &
        \multicolumn{1}{c|}{94.5} &
        \cellcolor[HTML]{D5E2F1}6.57 &
        \multicolumn{1}{c|}{94.1} &
        \cellcolor[HTML]{F3CECA}\textcolor{minired}{100} &
        \multicolumn{1}{c|}{89.6} &
        \cellcolor[HTML]{D5E2F1}14.0 &
        94.3 &
        \cellcolor[HTML]{F3CECA}\textcolor{minired}{100} &
        \multicolumn{1}{c|}{94.7} &
        \cellcolor[HTML]{D5E2F1}0.86 &
        \multicolumn{1}{c|}{94.4} &
        \cellcolor[HTML]{F3CECA}99.8 &
        \textcolor{minired}{94.9} 
        & 50.9 & 93.8 & 100 & \multicolumn{1}{|c}{89.6} \\ \hline
        
    \multicolumn{1}{c|}{\textbf{Beatrix}} &
        \cellcolor[HTML]{F3CECA}55.6 &
        \multicolumn{1}{c|}{93.8} &
        \cellcolor[HTML]{F3CECA}\textcolor{minired}{94.9} &
        \multicolumn{1}{c|}{93.8} &
        \cellcolor[HTML]{D5E2F1}2.13 &
        \multicolumn{1}{c|}{94.1} &
        \cellcolor[HTML]{D5E2F1}17.0 &
        94.2 &
        \cellcolor[HTML]{D5E2F1}4.12 &
        \multicolumn{1}{c|}{\textcolor{minired}{94.8}} &
        \cellcolor[HTML]{D5E2F1}8.64 &
        \multicolumn{1}{c|}{94.3} &
        \cellcolor[HTML]{F3CECA}90.4 &
        94.5 
        &39.0 & 94.2 & 94.9 & \multicolumn{1}{|c}{93.8} \\ \hline
        
    \multicolumn{1}{c|}{\textbf{AC}} &
        \cellcolor[HTML]{F3CECA}81.3 &
        \multicolumn{1}{c|}{76.9} &
        \cellcolor[HTML]{F3CECA}93.3 &
        \multicolumn{1}{c|}{82.1} &

        \cellcolor[HTML]{F3CECA}99.7 &
        \multicolumn{1}{c|}{83.1} &
        \cellcolor[HTML]{F3CECA}83.5 &
        81.3 &
        \cellcolor[HTML]{D5E2F1}4.31 &
        \multicolumn{1}{c|}{\textcolor{minired}{94.8}} &
        \cellcolor[HTML]{D5E2F1}7.63 &
        \multicolumn{1}{c|}{87.7} &
        \cellcolor[HTML]{F3CECA}\textcolor{minired}{100} &
        90.7 
        & 67.1& 85.0 & 100 & \multicolumn{1}{|c}{76.9} \\ \hline

    \multicolumn{1}{c|}{\textbf{ABL}} &
        \cellcolor[HTML]{F3CECA}88.6 &
        \multicolumn{1}{c|}{92.5} &
        \cellcolor[HTML]{F3CECA}94.2 &
        \multicolumn{1}{c|}{\textcolor{minired}{88.7}} &
        \cellcolor[HTML]{F3CECA}90.2 &
        \multicolumn{1}{c|}{93.1} &
        \cellcolor[HTML]{F3CECA}30.6 &
        94.2 &
        \cellcolor[HTML]{D5E2F1}6.32 &
        \multicolumn{1}{c|}{94.7} &
        \cellcolor[HTML]{D5E2F1}7.63 &
        \multicolumn{1}{c|}{94.4} &
        \cellcolor[HTML]{F3CECA}\textcolor{minired}{99.3} &
        94.9 
        & 59.6 & 93.2 & 99.3 & \multicolumn{1}{|c}{88.7}  \\ \hline

    \multicolumn{1}{c|}{\textbf{Strip}} &
        \cellcolor[HTML]{F3CECA}76.9 &
        \multicolumn{1}{c|}{\textcolor{minired}{85.3}} &
        \cellcolor[HTML]{F3CECA}93.8 &
        \multicolumn{1}{c|}{87.1} &
        \cellcolor[HTML]{F3CECA}98.6 &
        \multicolumn{1}{c|}{91.7} &
        \cellcolor[HTML]{F3CECA}25.5 &
        91.0 &
        \cellcolor[HTML]{D5E2F1}0.38 &
        \multicolumn{1}{c|}{94.8} &
        \cellcolor[HTML]{D5E2F1}9.63 &
        \multicolumn{1}{c|}{94.4} &
        \cellcolor[HTML]{F3CECA}\textcolor{minired}{99.8} &
        94.9 
        & 57.8 & 91.3 & 99.8 & \multicolumn{1}{|c}{81.3} \\ \hline

    \multicolumn{1}{c|}{\textbf{CT}} &
        \cellcolor[HTML]{D5E2F1}3.42 &
        \multicolumn{1}{c|}{93.1} &
        \cellcolor[HTML]{F3CECA}31.3 &
        \multicolumn{1}{c|}{91.2} &
        \cellcolor[HTML]{D5E2F1}0.53 &
        \multicolumn{1}{c|}{92.5} &
        \cellcolor[HTML]{D5E2F1}1.12 &
        93.2 &
        \cellcolor[HTML]{D5E2F1}0.44 &
        \multicolumn{1}{c|}{\textcolor{minired}{91.1}} &
        \cellcolor[HTML]{D5E2F1}2.16 &
        \multicolumn{1}{c|}{93.2} &
        \cellcolor[HTML]{F3CECA}\textcolor{minired}{100} &
        94.1
        & 19.9 & 92.6 & 100 & \multicolumn{1}{|c}{91.1} \\ \hline

    \multicolumn{1}{c|}{\textbf{Ours}} &
        \cellcolor[HTML]{D5E2F1}2.68 &
        \multicolumn{1}{c|}{94.9} &
        \cellcolor[HTML]{D5E2F1}0.44 &
        \multicolumn{1}{c|}{95.2} &
        \cellcolor[HTML]{D5E2F1}1.89 &
        \multicolumn{1}{c|}{\textcolor{minired}{93.1}} &
        \cellcolor[HTML]{D5E2F1}1.55 &
        94.8 &
        \cellcolor[HTML]{D5E2F1}1.16 &
        \multicolumn{1}{c|}{94.9} &
        \cellcolor[HTML]{D5E2F1}1.14 &
        \multicolumn{1}{c|}{94.4} &
        \cellcolor[HTML]{D5E2F1}\textcolor{minired}{9.68} &
        94.9
        & \textbf{2.65} & \textbf{94.6} & \textbf{9.68} & \multicolumn{1}{|c}{\textbf{93.1}} \\ \hline
    \end{tabular}}
    \caption{(\textcolor{blue}{a}) Upstream and (\textcolor{blue}{b}) Downstream evaluation and comparison results under \textbf{\underline{Case-0}}, CIFAR-10. 
    We list the poison ratio of each attack at the top of each column, which follows the original work that proposed these attacks. 
    We highlight the ASR below 20\% in \scalebox{0.9}{\colorbox[HTML]{D5E2F1}{\textbf{blue}}} as a success defense, the ASR above 20\% in \scalebox{0.9}{\colorbox[HTML]{F3CECA}{\textbf{red}}} as a failed defense case.
    }
    \label{tab:SLcifar10}
    \vspace{-1.5em}
    \end{table*}

\noindent
\textbf{Dataset \& models.}
We incorporate three standard computer vision benchmark datasets into our evaluation: CIFAR-10 \cite{krizhevsky2009learning} (main text), STL-10 \cite{coates2011analysis} (Appendix \ref{sec:newres}), and ImageNet \cite{deng2009imagenet} (a randomly selected 100-class subset, Appendix \ref{sec:newres}). To ensure the effectiveness of the baselines and fair comparison, we set the base set size as 1000 for all the settings. We will later show that our method is robust to different choices of the base set size in the ablation study, Appendix \ref{sec:ablation}. We obtain a 1000-size clean base set for each dataset by randomly selecting the samples from the test set and removing their label information. 
All the upstream evaluation metrics (i.e., TPR and FPR) are evaluated on the respective training sets, i.e., the training set of \textbf{\underline{Case-0}}, the fine-tuning set of \textbf{\underline{Case-2}}, and  the unlabeled pre-training set for \textbf{\underline{Case-1}}.
For \textbf{\underline{Case-0}}, we adopt all the remaining data from the test set for evaluation of the downstream metrics (i.e., ACC and ASR). For \textbf{\underline{Case-1}} and \textbf{\underline{Case-2}}, we split the remaining test set into half being fine-tuning set and half being the downstream metric evaluation set. 
ResNet-18 \cite{he2016deep} is adopted on the CIFAR-10. 
ViT-Small/16 \cite{dosovitskiy2020image} is adopted on STL-10 and ImageNet (Appendix \ref{sec:newres}). 
For \textbf{\underline{Case-1}}, we incorporate four state-of-the-art SSL training methods, i.e., SimCLR \cite{chen2020simple}, MoCo V3 \cite{chen2021empirical}, BYOL \cite{grill2020bootstrap}, and the MAE \cite{he2022masked}, for evaluation. For \textbf{\underline{Case-2}}, we consider the two most popular transfer learning cases, namely, \textsf{FT-all} and \textsf{FT-last} (detailed in Section~\ref{sec:background}).
The pre-trained model parameters for fine-tuning are loaded from the timm library\footnote{\url{https://timm.fast.ai/}}.


\noindent
\textbf{Baseline defenses.}
Referring to Table \ref{tab:compare}, we incorporate a wide range of existing backdoor detection for comparison, including both standard baselines used in prior work as well as state-of-the-art ones. 
In particular, we consider Spectral
\cite{tran2018spectral}, Spectre
\cite{hayase2021spectre}, and the Beatrix
\cite{ma2022beatrix}; 
we include AC
\cite{chen2018detecting} as a representative work that utilizes intermediate neural activation; 
ABL
\cite{li2021anti}, which was originally a robust training defense and repurposed as a detection method based on output losses; 
Strip
\cite{gao2019strip}
as a representative detection approach based on model outputs;
and CT
\cite{qi2022fight}, the most recent work reported achieving state-of-the-art performance on end-to-end SL settings based on confusion training.
All the implementations and hyperparameters follow the original papers. For methods that rely on or can be boosted by an additional base set, e.g., Spectre, Beatrix, Strip, CT, we use the same 1000-size base set as ours. We note that this comparison setting might not be fair, as compared to these baselines, our method relaxes the requirement on label information; in addition, AC and ABL cannot be adapted to use the base set.
We want to show that even without label information,
our method can still achieve comparable or much better results with stronger robustness than the other baselines.
Detailed explanations of the defense settings and how we adapted them to \textbf{\underline{Case-1}} and \textbf{\underline{Case-2}} are provided in Appendix \ref{sec:defense_set}.

\noindent
\textbf{Backdoor attack settings.}
For \textbf{\underline{Case-0}} we incorporate seven standard or state-of-the-art attacks, including four dirty-label and clean-label ones. 
For dirty-label backdoor attacks, we incorporate localized backdoor attack BadNets \cite{gu2017badnets}, global-wised blended trigger Blended \cite{chen2017targeted}, wrapping-based invisible backdoor attack WaNet \cite{nguyen2021wanet}, and the state-of-the-art sample-specific invisible backdoor attack, ISSBA \cite{li2021invisible}. For clean-label attacks, we include the standard Label Consistent (LC) attack \cite{turner2019label}, the state-of-the-art feature-collision-based hidden trigger backdoor, Sleeper Agent Attack (SAA) \cite{souri2021sleeper}, and the state-of-the-art optimization-based Narcissus attack (Narci.) \cite{zeng2022narcissus}.
For \textbf{\underline{Case-1}}, only limited existing work has explored the attack over SSL's unlabeled training set. We incorporate
the Checkerboard trigger (C-brd) used in \cite{carlini2021poisoning}, the Colored Square trigger (C-squ) used in \cite{saha2022backdoor}, and the state-of-the-art YCbCr frequency-based invisible trigger used in CTRL \cite{li2022demystifying}. In particular, CTRL has been shown to achieve a magnitude higher attacking efficacy than \cite{saha2022backdoor}.
For \textbf{\underline{Case-2}}, directly implementing some of the attacks from end-to-end SL may not lead to effective attacks, e.g., the Blended attack cannot achieve high ASR under the \textsf{FT-all} settings. Thus, we consider attacks that can maintain effectiveness for each TL setting.
BadNets and the SAA are adopted for evaluation under the \textsf{FT-all} case. Blended and the hidden trigger backdoor attack (HTBA) \cite{saha2020hidden} are adopted for the evaluation under the \textsf{FT-last} case. All the incorporated attacks' settings, such as trigger design and trigger strength, all follow their original papers.
%
Appendix \ref{sec:attack_set} details the specifics of these attacks' setups under each learning paradigm and visual examples of the poisoned samples we intend to detect.



\renewcommand{\arraystretch}{1.3}
\begin{table}[t!]
\vspace{-.5em}
\centering
\resizebox{\columnwidth}{!}{
\begin{tabular}{c|c||P{1.2cm}|P{1.2cm}|P{1.2cm}|P{1.2cm}|P{1.2cm}|P{1.2cm}|P{1.2cm}|P{1.2cm}} 
\hline
       & \begin{tabular}[c]{@{}c@{}}Poison\\Ratio\%\end{tabular} & \begin{tabular}[c]{@{}c@{}}\textbf{Spectral}\\\cite{tran2018spectral}\end{tabular} & \begin{tabular}[c]{@{}c@{}}\textbf{Spectre}\\\cite{hayase2021spectre}\end{tabular} & \begin{tabular}[c]{@{}c@{}}\textbf{Beatrix}\\\cite{ma2022beatrix}\end{tabular} & \begin{tabular}[c]{@{}c@{}}\textbf{AC}\\\cite{chen2018detecting}\end{tabular} & \begin{tabular}[c]{@{}c@{}}\textbf{ABL}\\\cite{ma2022beatrix}\end{tabular} & \begin{tabular}[c]{@{}c@{}}\textbf{Strip}\\\cite{gao2019strip}\end{tabular} & \begin{tabular}[c]{@{}c@{}}\textbf{CT}\\\cite{ma2022beatrix}\end{tabular} & \textbf{Ours}  \\ 
\hline
\hline
\multicolumn{1}{c|}{} & 0.05\%$\backslash$25 & 25  & 23  & 13             & 22    & 7      & 19                                                         & 1            & \textbf{0}              \\ 
\cline{2-10}
\multicolumn{1}{c|}{} & 1\%$\backslash$500 & \cellcolor[HTML]{F3CECA}37  & 23  & 13             & \cellcolor[HTML]{F3CECA}416    & \cellcolor[HTML]{F3CECA}32      & \cellcolor[HTML]{F3CECA}446                                                         & \textbf{0}            & 17              \\ 
\cline{2-10}
\multicolumn{1}{c|}{} & 5\%$\backslash$2500 & \cellcolor[HTML]{F3CECA}109    & \cellcolor[HTML]{F3CECA}109    & \cellcolor[HTML]{F3CECA}155        & \cellcolor[HTML]{F3CECA}238     & \cellcolor[HTML]{F3CECA}365     & \cellcolor[HTML]{F3CECA}1866                                                     & 20           & \textbf{13}             \\ 
\cline{2-10}
\multicolumn{1}{c|}{} & 20\%$\backslash$10000 & \cellcolor[HTML]{F3CECA}817    & \cellcolor[HTML]{F3CECA}170  & \cellcolor[HTML]{F3CECA}113          & \cellcolor[HTML]{F3CECA}1086    & \cellcolor[HTML]{F3CECA}4590     & \cellcolor[HTML]{F3CECA}330                                                   & 16           & \textbf{7}              \\
\cline{2-10}
\multicolumn{1}{c|}{\multirow{-5}{*}{\textbf{\begin{tabular}[c]{@{}c@{}}\rotatebox{90}{BadNets }\end{tabular}}}} & 50\%$\backslash$25000 & \cellcolor[HTML]{F3CECA}7963    & \cellcolor[HTML]{F3CECA}158  & \cellcolor[HTML]{F3CECA}264          & \cellcolor[HTML]{F3CECA}774    & \cellcolor[HTML]{F3CECA}1944     & \cellcolor[HTML]{F3CECA}1001                                                   & \cellcolor[HTML]{F3CECA}25000            & \textbf{4}              \\
\hline
\hline
\multicolumn{1}{c|}{} & 0.05\%$\backslash$25 & 25  & 25  & 22             & 25    & 19      & 23                                                         & \textbf{2}            & 4              \\ 
\cline{2-10}
\multicolumn{1}{c|}{}  & 1\%$\backslash$500 & \cellcolor[HTML]{F3CECA}86 & \cellcolor[HTML]{F3CECA}44 & \cellcolor[HTML]{F3CECA}53 & \cellcolor[HTML]{F3CECA}49          & \cellcolor[HTML]{F3CECA}41        & \cellcolor[HTML]{F3CECA}413                                       &    16                 & \textbf{2}                     \\ 
\cline{2-10}
\multicolumn{1}{c|}{}  & 5\%$\backslash$2500 & 6 & 5 & \cellcolor[HTML]{F3CECA}803 & \cellcolor[HTML]{F3CECA}866           & \cellcolor[HTML]{F3CECA}1023       & \cellcolor[HTML]{F3CECA}2068                                       &    \cellcolor[HTML]{F3CECA}33                 & \textbf{0}                     \\ 
\cline{2-10}
\multicolumn{1}{c|}{}  & 20\%$\backslash$10000 & \cellcolor[HTML]{F3CECA}226 & 27 & \cellcolor[HTML]{F3CECA}31 & \cellcolor[HTML]{F3CECA}306        & \cellcolor[HTML]{F3CECA}4965          & \cellcolor[HTML]{F3CECA}1669                                       &    \cellcolor[HTML]{F3CECA}3659                 & \textbf{13}                     \\ \cline{2-10}
\multicolumn{1}{c|}{\multirow{-5}{*}{\textbf{\begin{tabular}[c]{@{}c@{}}\rotatebox{90}{Blended }\end{tabular}}}} & 50\%$\backslash$25000 & \cellcolor[HTML]{F3CECA}9568    & \cellcolor[HTML]{F3CECA}1023  & \cellcolor[HTML]{F3CECA}2386          & \cellcolor[HTML]{F3CECA}1514    & \cellcolor[HTML]{F3CECA}13959     & \cellcolor[HTML]{F3CECA}10736                                                   & \cellcolor[HTML]{F3CECA}25000            & \textbf{8}              \\
\hline
\end{tabular}}
\vspace{-.5em}
\caption{\# poisons remained in the filtered training set after defense (\textbf{\underline{Case-0}}, CIFAR-10). 
\textbf{Bolded} results denote the smallest value. \scalebox{0.9}{\colorbox[HTML]{F3CECA}{red}} to highlight failed defenses where more than 30 poisoned samples remain as we find this amount of poisons still enables ASRs greater than 30\%.}
\label{tab:ratioablation}
\vspace{-1.5em}
\end{table}

\vspace{-1.2em}
\subsection{\underline{\textbf{Case-0}}: End-to-end SL}
\label{sec:case0exp}
\vspace{-.8em}

\noindent
\textbf{Detection performance against different attacks in SL.}
Table \ref{tab:SLcifar10} presents the upstream and downstream evaluation results under the end-to-end SL setting on the CIFAR-10 dataset with the ResNet-18 model trained from scratch for 200 epochs. For each different attack, we adopt the poison ratio following each original paper, which is listed at the top of each column. We have included the row of ``No Defense'' in Table \ref{tab:SLcifar10} (\textcolor{blue}{b}) to show the attack effects without any backdoor detection defense in place.
%
Existing methods are able to achieve decent detection effects on some specific attacks, but they experience large performance variations when defending different attacks. 
These methods either solely rely on the embedding space of a poisoned model that may change with different trigger designs or rely on some detection rule that may not apply to specific backdoor designs.
For example, ABL assumes that backdoor samples achieve the lowest loss at the early stage of training. However, the Narci. clean-label poisoned samples' losses do not meet the assumption; thus, ABL is not effective on the Narci. 
The recently proposed CT achieves the highest detection rate and the most consistent performance among all baselines, but it still fails to detect the state-of-the-art clean-label attack, Narci. Notably, no existing detection method obtains satisfying results as Narci. introduces optimized features as robust as the semantic features of the target class \cite{zeng2022narcissus}.
Regarding the upstream evaluation in Table \ref{tab:SLcifar10} (\textcolor{blue}{a}), our method reliably achieves a TPR above 90\% for all the evaluated settings and significantly improves the state-of-the-art in terms of the average and worse-case defensive performance over different attacks.
%
%
Regarding the downstream evaluation in Table \ref{tab:SLcifar10} (\textcolor{blue}{b}), we find that $\AlgName$ is the only defense that gives rise to robust models over all the evaluated poisoned datasets, i.e., all ASRs drop below random guessing rate, i.e., 10\%.
In particular, our method is the only effective method to mitigate Narci. 
Moreover, the downstream models trained over $\AlgName$ filtered datasets achieve the highest average ACC. Notably, the average ACC of our method is slightly higher than using the original poisoned dataset (which contains more clean samples).
Results for multiple attacks introduced simultaneously are provided in Appendix \ref{sec:newres}, with similar observations.

\begin{figure}[t!]
  \centering
     \vspace{-1.5em}
  \includegraphics[width=0.66\linewidth]{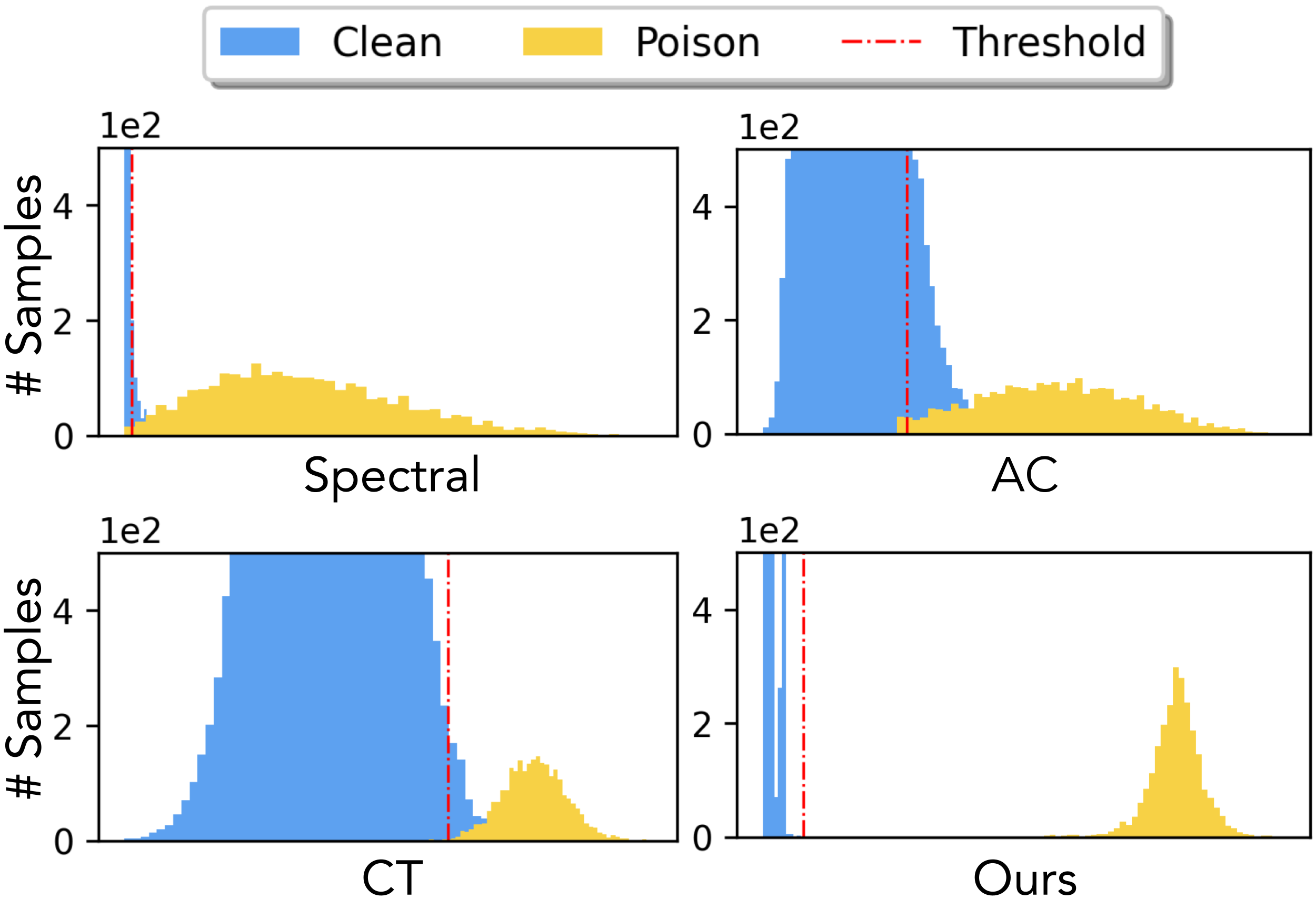}
  \vspace{-1.em}
  \caption{
 Detection results with different defenses in distribution histograms (CIFAR-10, Blended attack, 5\%, \textbf{\underline{Case-0}}). We emphasize the effects and the necessity of adaptive thresholding and the process of actively pushing the distribution of clean and poison away from each other.
  }
  \label{fig:slcompare}
  \vspace{-1.5em}
\end{figure}

Unlike $\AlgName$, the existing methods do not have an active process to induce differentiating behaviors between clean samples and poisoned ones. Thus clean and poisoned samples often have overlapping behaviors and cannot be easily separated.
We illustrate the separation between clean and poisoned samples using different detection methods and their threshold in Figure \ref{fig:slcompare}, emphasizing the importance of the proposed active offset process.

\noindent
\textbf{Impact of poison ratios.}
%
In Table \ref{tab:ratioablation}, we study the effects of poison ratio on different detection methods against two standard attacks, namely, BadNets, and the Blended attack. 
Most existing detection works better for small poison ratios but fails as the ratio increases. 
One reason is that many works, such as Spectral, Spectre, and AC, are based on the feature distribution of the poison dataset. However, an increased poisoning rate will cause the clean feature distribution to be closer to the poisoned one, making them less separable.
CT is the most robust baseline in the previous evaluation, but it also fails for very large ratios like 20\% (10000 poisons) or 50\% (25000 poisons). The reason could be that their detector uses fixed hyperparameters that are fine-tuned on small poison ratios.
Our defense is robust to poison ratio changes, even for extreme cases where half of the samples in the training set are poisoned or only 25 ($0.05\%$) samples are poisoned. 
%

\vspace{-1.em}
\subsection{\underline{\textbf{Case-1}}: SSL Adaptation}
\label{sec:sslexp}

\begin{table}[h!]
\centering
       \vspace{-1.0em}
      \resizebox{\columnwidth}{!}{
      \begin{tabular}{
      >{\columncolor[HTML]{FFFFFF}}c |
      >{\columncolor[HTML]{FFFFFF}}c 
      >{\columncolor[HTML]{FFFFFF}}c |
      >{\columncolor[HTML]{FFFFFF}}c 
      >{\columncolor[HTML]{FFFFFF}}c |
      >{\columncolor[HTML]{FFFFFF}}c 
      >{\columncolor[HTML]{FFFFFF}}c  ||
      cc||cc}
      \hline
      \cellcolor[HTML]{FFFFFF}{\color[HTML]{000000} } &
          \multicolumn{2}{c|}{\textbf{C-brd (0.5\%)}} &
  \multicolumn{2}{c|}{\textbf{C-Squ (0.5\%)}} &
  \multicolumn{2}{c||}{\textbf{ CTRL (1\%)}} &
  \multicolumn{2}{c||}{\textbf{Average}}&
  \multicolumn{2}{c}{\textbf{Worst-Case}}\\ \cline{2-11} 
      &
        {\color[HTML]{000000} TPR}  $\uparrow$&
        {\color[HTML]{000000} FPR}  $\downarrow$&
        {\color[HTML]{000000} TPR}  $\uparrow$&
        {\color[HTML]{000000} FPR}  $\downarrow$&
        {\color[HTML]{000000} TPR}  $\uparrow$&
        {\color[HTML]{000000} FPR}  $\downarrow$&
        TPR  $\uparrow$&
        FPR $\downarrow$&
        TPR  $\uparrow$&
        \multicolumn{1}{|c}{FPR $\downarrow$}\\ \hline
      \textbf{Spectral} &
        {\color[HTML]{000000} \textcolor{minired}{0.08}} &
        {\color[HTML]{000000} \textcolor{minired}{7.89}} &
        {\color[HTML]{000000} 0.44} &
        {\color[HTML]{000000} 7.87} &
        {\color[HTML]{000000} 1.20} &
        {\color[HTML]{000000} 1.50} &
        0.57 & 5.75 &0.08&\multicolumn{1}{|c}{7.89}\\ \hline
      \textbf{Spectre} &
         0.64 &
         \textcolor{minired}{7.86} &
         \textcolor{minired}{2.36} &
         7.77 &
         0.40 &
         1.51 &
         0.64 & 5.71 &0.40&\multicolumn{1}{|c}{7.86}\\ \hline
        {\color[HTML]{000000}\textbf{Beatrix}} & \textcolor{minired}{0.88} & \textcolor{minired}{7.85} & 2.76 & 7.75 & 73.4 & 2.79 & 25.7 & 6.13 &0.88&\multicolumn{1}{|c}{7.85}\\ \hline
        {\color[HTML]{000000}\textbf{AC}}  & \textcolor{minired}{6.72} & \textcolor{minired}{28.6} & 9.88 & 28.4 & 36.8 & 21.3 & 17.8 & 26.1 &6.72&\multicolumn{1}{|c}{28.6}\\ \hline
      
        \textbf{ABL} & \textcolor{minired}{5.76} & \textcolor{minired}{7.59} & 6.24 & 7.57 & 20.2 & 1.31 & 10.7 & 5.49&5.76&\multicolumn{1}{|c}{7.59}\\ \hline
      
      {\color[HTML]{000000} \textbf{Ours}} &
        {\color[HTML]{000000} \textcolor{minired}{91.5}} &
        {\color[HTML]{000000} 0.67} &
        {\color[HTML]{000000} 96.0} &
        {\color[HTML]{000000} 0.25} &
        {\color[HTML]{000000} 97.4} &
        {\color[HTML]{000000} \textcolor{minired}{0.69}} &
        \textbf{95.0} & \textbf{0.54}&\textbf{91.5}&\multicolumn{1}{|c}{\textbf{0.69}} \\ \hline
        
      \end{tabular}}
      \vspace{-0.5em}
      \caption{
      Upstream evaluation and comparison results under \textbf{\underline{Case-1}} with SimCLR. The \textbf{bolded} results denote the best defense results among the evaluated defenses.
      }
      \label{tab:SSLdefense}
      \vspace{-0.5em}
\end{table}

\begin{table}[t!]
\centering
        \vspace{-1.5em}
        \resizebox{0.75\columnwidth}{!}{
        \begin{tabular}{
        >{\columncolor[HTML]{FFFFFF}}c |
        >{\columncolor[HTML]{FFFFFF}}c 
        >{\columncolor[HTML]{FFFFFF}}c |
        >{\columncolor[HTML]{FFFFFF}}c 
        >{\columncolor[HTML]{FFFFFF}}c |
        >{\columncolor[HTML]{F3CECA}}c 
        >{\columncolor[HTML]{FFFFFF}}c }
        \hline
        \cellcolor[HTML]{FFFFFF}{\color[HTML]{000000} } &
          \multicolumn{2}{c|}{\textbf{C-brd (0.5\%)}} &
  \multicolumn{2}{c|}{\textbf{C-Squ (0.5\%)}} &
  \multicolumn{2}{c}{\textbf{ CTRL (1\%)}} \\ \cline{2-7} 
        &
          \cellcolor[HTML]{FFFFFF}{\color[HTML]{000000} ASR*} $\downarrow$&
          {\color[HTML]{000000} ACC} $\uparrow$&
          \cellcolor[HTML]{FFFFFF}{\color[HTML]{000000} ASR*} $\downarrow$&
          {\color[HTML]{000000} ACC} $\uparrow$&
          \cellcolor[HTML]{FFFFFF}{\color[HTML]{000000} ASR} $\downarrow$&
          {\color[HTML]{000000} ACC} $\uparrow$\\ \hline
        \textbf{No Def.} &
           404 &
           85.2 &
           435 &
           84.6 &
           81.4 &
           85.3 \\

           \hline
           \hline
        \textbf{Spectral}&
          {\color[HTML]{000000} 405} &
          {\color[HTML]{000000} 84.1} &
          {\color[HTML]{000000} 478} &
          {\color[HTML]{000000} 84.2} &
          {\color[HTML]{000000} 81.3} &
          {\color[HTML]{000000} 85.2} \\ \hline
        \textbf{Spectre} &
           405 &
           84.1 &
           445 &
           84.2 &
           81.4 &
           85.3 \\ \hline
          {\color[HTML]{000000}\textbf{Beatrix}} & 402 & 84.2 & 444 & 84.2 & \cellcolor[HTML]{D5E2F1}16.8 & 85.0\\ \hline
          {\color[HTML]{000000}\textbf{AC}}  & 513 & 73.26 & 376 & 73.2 & \cellcolor[HTML]{F3CECA}36.5 & 78.6 \\ \hline
        
          \textbf{ABL} & 380 & 84.6  & 399 & 84.4 & \cellcolor[HTML]{F3CECA}46.6 & 85.3 \\ \hline
        
        {\color[HTML]{000000} \textbf{Ours}} &
          \cellcolor[HTML]{FFFFFF}{{\color[HTML]{000000} 100}} &
          {\color[HTML]{000000} 85.1} &
          \cellcolor[HTML]{FFFFFF}{{\color[HTML]{000000} 87.0}} &
          {\color[HTML]{000000} 84.9} &
          \cellcolor[HTML]{D5E2F1}{\color[HTML]{000000} 2.47} &
          {\color[HTML]{000000} 85.9} \\ \hline
          
        \end{tabular}}
        \caption{
        Downstream evaluation and comparison results under \textbf{\underline{Case-1}} with SimCLR. 
        We highlight the ASR below 20\% in \scalebox{0.9}{\colorbox[HTML]{D5E2F1}{\textbf{blue}}} as a success defense, the ASR above 20\% in \scalebox{0.9}{\colorbox[HTML]{F3CECA}{\textbf{red}}} as a failed defense case. 
        ASR$^*$ is the number of successfully attacked samples. We use ASR$^*$ instead for the C-brd and the C-Squ attack, referring to the original work \cite{saha2022backdoor}, as their ASRs are naturally low to SSL paradigms.
        }
        \label{tab:SSLdefensedown}
        \vspace{-.5em}
\end{table}

\noindent
\textbf{Detection performance against different attacks in SSL.}
Now we study the efficacy in detecting unlabeled poisons under the SSL adaptation cases. 
Table \ref{tab:SSLdefense} and Table \ref{tab:SSLdefensedown} list out the upstream and downstream evaluation results, respectively, on CIFAR-10 using ResNet-18 trained via SimCLR-based SSL for 600 epochs with linear adaptation for 100 epochs. We find that the ASRs of C-brd and C-Squ are below 20\% so these attacks cannot lead to a successful attack on average. We still keep their results but show the number of successfully attacked samples (denoted with $ASR^*$) as done in \cite{saha2022backdoor}. 
Even though these attacks do not result in as high ASR as the attacks in SL or as the CTRL attack, they can still result in an increase of samples with triggers being classified as the target class. 
As shown in Table \ref{tab:SSLdefense}, among all the evaluated attacks, our method obtains the highest TRP while remaining the lowest FPR among all detection methods.
Noting the absence of CT under the SSL. Recall that in the SL setting, CT can achieve compatible results as our method on most attack settings; yet, it is inapplicable to SSL as its core technique---confusion training---relies on label information~\cite{qi2022fight}. 
In particular, as C-brd and C-Squ do not result in a high ASR as shown in Table \ref{tab:SSLdefensedown}, the model's response to clean and backdoor samples is not sufficiently different, thereby making detection very difficult. In fact, none of the baselines provides reliable detection of these two attacks. For the CTRL attack, which achieves an ASR of over 80\%, we start to see that some of the baseline defenses take effect, e.g., the Beatrix. But still, our method achieves the best upstream detection performance (Table \ref{tab:SSLdefense}) and gives rise to the highest ACC and lowest ASR downstream (Table \ref{tab:SSLdefensedown}).


\noindent
\textbf{Further evaluation with more SSL training algorithms.}
We further evaluate our defense under other popular SSL training algorithms and different model structures and datasets, e.g.,  ResNet-18 and ViT-Small/16 trained using SimCLR, MoCO V3, BYOL, MAE over CIFAR-10 or the ImageNet (Appendix \ref{sec:newres}). The upstream and downstream evaluation results on the CIFAR-10 are shown in TBALE \ref{tab:SSLcifar10} and Table \ref{tab:SSLcifar10down}, respectively. 
%
Across all the evaluated settings, our method provides reliable upstream detection results with TPRs over 90\% for all the cases and low FPRs. Thanks to the upstream efficacy, our detection method can give rise to the downstream model with a low ASR and an ACC close to or better than the settings without removing any training point. 
Overall, our results demonstrate that our method can reliably sift out the poisoned samples across different settings of SSL adaptation.

\begin{table}[t!]
\centering
\vspace{-.5em}
\resizebox{0.7\columnwidth}{!}{
\begin{tabular}{ccc|cc|cc}
\hline
       \multicolumn{1}{c|}{} &
          \multicolumn{2}{c|}{\textbf{C-brd (0.5\%)}} &
  \multicolumn{2}{c|}{\textbf{C-Squ (0.5\%)}} &
  \multicolumn{2}{c}{\textbf{ CTRL (1\%)}} \\ 
     \hhline{~|-|-|-|-|-|-|}
  \multicolumn{1}{c|}{} &
  TPR $\uparrow$&
  FPR $\downarrow$&
  TPR $\uparrow$&
  FPR $\downarrow$&
  TPR $\uparrow$&
  FPR $\downarrow$\\
  \hline
\multicolumn{1}{c|}{\textbf{SimCLR}} &
  \textcolor{black}{91.5} &
  0.67 &
  96.0 &
  0.25 &
  97.4 &
  \textcolor{black}{0.69} \\ \hline
\multicolumn{1}{c|}{\textbf{MoCo V3}} &
  \textcolor{black}{91.3} &
  \textcolor{black}{0.49} &
  96.9 &
  0.20 &
  98.2 &
  0.32 \\ \hline
\multicolumn{1}{c|}{\textbf{BYOL}} &
  95.9 &
  0.22 &
  95.8 &
  0.35 &
  \textcolor{black}{94.6} &
  \textcolor{black}{0.57} \\ \hline
\multicolumn{1}{c|}{\textbf{MAE}} &
  \textcolor{black}{97.2} &
  0.67 &
  98.2 &
  0.50 &
  \textcolor{black}{97.2} &
  \textcolor{black}{0.73} \\ \hline
\end{tabular}}
\caption{
Further upstream evaluation of $\AlgName$ under \textbf{\underline{Case-1}} with four SSL training algorithms, CIFAR-10. 
}
\label{tab:SSLcifar10}
\vspace{-1.5em}
\end{table}

\noindent
\textbf{Impact of \# logits w.r.t. SSL downstream task.}
Note that for SSL evaluation, the pre-trained model requires a fixed number of logits, each corresponding to a different output category. In our evaluation, we use the actual classes contained (e.g., 10 for the CIFAR-10 and 100 for the ImageNet 100-subset). 
Such a setting is applicable when the defender knows the exact downstream classification task. Now we consider a much more strict case where one tries to conduct detection over unlabeled datasets without any prior knowledge about the 
number of categories in downstream tasks.
As shown in Table \ref{tab:logitsnumber}, we find our method is robust to the change in the number of logits and can maintain a TPR higher than 90\%. 

\begin{table*}[t!]
\centering
\vspace{-1.5em}
\resizebox{0.8\textwidth}{!}{
\begin{tabular}{
>{\columncolor[HTML]{FFFFFF}}c 
>{\columncolor[HTML]{FFFFFF}}c 
>{\columncolor[HTML]{FFFFFF}}c 
>{\columncolor[HTML]{FFFFFF}}c ||
>{\columncolor[HTML]{FFFFFF}}c 
>{\columncolor[HTML]{FFFFFF}}c 
>{\columncolor[HTML]{FFFFFF}}c 
>{\columncolor[HTML]{FFFFFF}}c |
>{\columncolor[HTML]{FFFFFF}}c 
>{\columncolor[HTML]{FFFFFF}}c 
>{\columncolor[HTML]{FFFFFF}}c 
>{\columncolor[HTML]{FFFFFF}}c |
>{\columncolor[HTML]{FFFFFF}}c 
>{\columncolor[HTML]{FFFFFF}}c 
>{\columncolor[HTML]{FFFFFF}}c 
>{\columncolor[HTML]{FFFFFF}}c }
\hline

 \multicolumn{1}{c|}{} &
  \multicolumn{3}{c||}{\cellcolor[HTML]{FFFFFF}{\color[HTML]{000000} \textbf{No Attack}}} &
  \multicolumn{4}{c|}{\cellcolor[HTML]{FFFFFF}{\color[HTML]{000000} \textbf{C-brd (5\%)}}} &
  \multicolumn{4}{c|}{\cellcolor[HTML]{FFFFFF}{\color[HTML]{000000} \textbf{C-Squ (5\%)}}} &
  \multicolumn{4}{c}{\cellcolor[HTML]{FFFFFF}{\color[HTML]{000000} \textbf{CTRL (1\%)}}} \\ 
     \hhline{~|-|-|-|-|-|-|-|-|-|-|-|-|-|-|-|}
  \multicolumn{1}{c|}{} &
  {\color[HTML]{000000} ASR} $\downarrow$&
  {\color[HTML]{000000} ASR*} $\downarrow$&
  {\color[HTML]{000000} ACC} $\uparrow$&
  {\color[HTML]{000000} ASR*$_{0}$} &
  \cellcolor[HTML]{FFFFFF}{{\color[HTML]{000000} ASR*}} $\downarrow$&
  {\color[HTML]{000000} ACC$_{0}$} &
  {\color[HTML]{000000} ACC} $\uparrow$&
  {\color[HTML]{000000} ASR*$_{0}$} &
  \cellcolor[HTML]{FFFFFF}{{\color[HTML]{000000} ASR*}} $\downarrow$&
  {\color[HTML]{000000} ACC$_{0}$} &
  {\color[HTML]{000000} ACC} $\uparrow$&
  {\color[HTML]{000000} ASR$_{0}$} &
  \cellcolor[HTML]{FFFFFF}{{\color[HTML]{000000} ASR}} $\downarrow$&
  {\color[HTML]{000000} ACC$_{0}$} &
  {\color[HTML]{000000} ACC} $\uparrow$\\
  \hline
  
\multicolumn{1}{c|}{\textbf{SimCLR}} &
  {\color[HTML]{000000} 1.78} &
  {\color[HTML]{000000} 79} &
  {\color[HTML]{000000} 85.4} &
  {\color[HTML]{000000} 403} &
  {\color[HTML]{000000} 100} &
  {\color[HTML]{000000} 84.7} &
  {\color[HTML]{000000} 84.8} &
  {\color[HTML]{000000} 434} &
  {\color[HTML]{000000} 87} &
  {\color[HTML]{000000} 84.6} &
  {\color[HTML]{000000} 85.0} &
  {\color[HTML]{000000} 61.4} &
  {\color[HTML]{000000} 2.47} &
  {\color[HTML]{000000} 85.3} &
  {\color[HTML]{000000} 85.9} \\ \hline
\multicolumn{1}{c|}{\textbf{MoCo V3}} &
  {\color[HTML]{000000} 1.88} &
  {\color[HTML]{000000} 83} &
  {\color[HTML]{000000} 87.2} &
  {\color[HTML]{000000} 411} &
  {\color[HTML]{000000} 95} &
  {\color[HTML]{000000} 87.0} &
  {\color[HTML]{000000} 87.1} &
  {\color[HTML]{000000} 374} &
  {\color[HTML]{000000} 83} &
  {\color[HTML]{000000} 87.2} &
  {\color[HTML]{000000} 87.13} &
  {\color[HTML]{000000} 56.3} &
  {\color[HTML]{000000} 3.70} &
  {\color[HTML]{000000} 86.5} &
  {\color[HTML]{000000} 87.9} \\ \hline
\multicolumn{1}{c|}{\textbf{BYOL}} &
  {\color[HTML]{000000} 1.13} &
  {\color[HTML]{000000} 50} &
  {\color[HTML]{000000} 85.6} &
  {\color[HTML]{000000} 455} &
  {\color[HTML]{000000} 79} &
  {\color[HTML]{000000} 85.5} &
  {\color[HTML]{000000} 85.3} &
  {\color[HTML]{000000} 446} &
  {\color[HTML]{000000} 56} &
  {\color[HTML]{000000} 85.2} &
  {\color[HTML]{000000} 85.4} &
  {\color[HTML]{000000} 39.7} &
  {\color[HTML]{000000} 4.36} &
  {\color[HTML]{000000} 85.5} &
  {\color[HTML]{000000} 85.5} \\ \hline
\multicolumn{1}{c|}{\textbf{MAE}} &
  {\color[HTML]{000000} 1.58} &
  {\color[HTML]{000000} 70} &
  {\color[HTML]{000000} 89.2} &
  {\color[HTML]{000000} 83} &
  {\color[HTML]{000000} 74} &
  {\color[HTML]{000000} 88.4} &
  {\color[HTML]{000000} 88.4} &
  {\color[HTML]{000000} 104} &
  {\color[HTML]{000000} 70} &
  {\color[HTML]{000000} 88.65} &
  {\color[HTML]{000000} 88.93} &
  {\color[HTML]{000000} 15.9} &
  {\color[HTML]{000000} 3.42} &
  {\color[HTML]{000000} 87.2} &
  {\color[HTML]{000000} 89.9} \\ \hline
\end{tabular}}
\vspace{-.3em}
\caption{
Downstream evaluation results of our method under \textbf{\underline{Case-1}}, CIFAR-10.
ASR$^*$ is the number of successfully attacked samples. ASR*$_{0}$ and ACC$_{0}$ with subscripts are the results without defense (i.e., the ``No Defense'' baseline in other tables). We use ASR$^*$ instead of ASR for the C-brd and the C-Squ attack, referring to the original work \cite{saha2022backdoor}, as their ASRs are naturally low. 
}
\label{tab:SSLcifar10down}
\vspace{-1.5em}
\end{table*}

\begin{table}[t!]
\centering
\resizebox{\columnwidth}{!}{
\begin{tabular}{c|cc|cc|cc|cc}
\hline
\multirow{2}{*}{ } &
  \multicolumn{2}{c|}{\textbf{5}} &
  \multicolumn{2}{c|}{\textbf{10}} &
  \multicolumn{2}{c|}{\textbf{100}} &
  \multicolumn{2}{c}{\textbf{1000}} \\ \cline{2-9} 
 &
  TPR $\uparrow$&
  FPR $\downarrow$&
  TPR $\uparrow$&
  FPR $\downarrow$&
  TPR $\uparrow$&
  FPR $\downarrow$&
  TPR $\uparrow$&
  FPR $\downarrow$\\ \hline
\textbf{CTRL (1\%)} &
  93.2 &
  0.02 &
  97.4 &
  0.07 &
  95.2 &
  0.34 &
  92.6 &
  2.81 \\ \hline
\end{tabular}}
\vspace{-0.6em}
\caption{
\# logits used and the detection effects over unlabeled CTRL poisons (\textbf{\underline{Case-1}}, CIFAR-10, SimCLR, ResNet-18).
}
\label{tab:logitsnumber}
\vspace{-1.5em}
\end{table}

\vspace{-1.em}
\subsection{\underline{\textbf{Case-2}}: Transfer Learning}
\label{sec:finetuneexp}

\noindent
\textbf{Detection performance against different attacks in TL.}
We consider two of the most popular TL schemes for evaluation: \textsf{FT-all} and \textsf{FT-last} with models pre-trained on the ImageNet. 
All the existing backdoor defenses can be easily generalized to TL. However, none of them has empirically evaluated the backdoor detection efficacy under the TL settings in the prior literature, which leaves a gap to fill.

\begin{figure}[h!]
  \centering
     \vspace{-.3em}
  \includegraphics[width=0.85\linewidth]{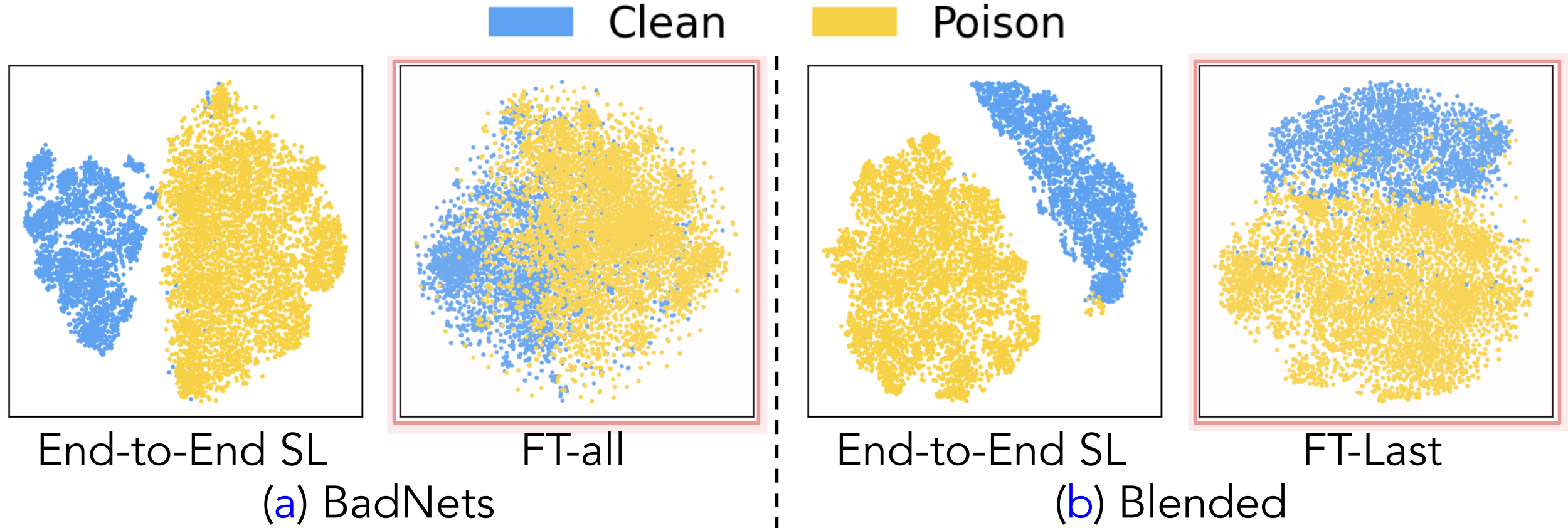}
  \vspace{-0.5em}
  \caption{
  In-class features space t-SNE results with the model trained with CIFAR-10 using end-to-end SL or TL: \textbf{(\textcolor{blue}{a})} BadNets 20\%, 
\textbf{(\textcolor{blue}{b})} Blended 20\%. 
  }
  \label{fig:tsne}
  \vspace{-.8em}
\end{figure}

The upstream and downstream results are listed in Table \ref{tab:TLcifar10}. 
Existing methods' detection results on \textsf{FT-all} seem more consistent than the results on \textsf{FT-last}. This observation might be due to that \textsf{FT-all} is a setting much closer to the end-to-end SL. While many defenses can achieve satisfying results on some specific attacks in SL, none can achieve a TPR above 90\% for all attack settings in TL, except CT on BadNets. 
We now take a closer look at the reason why existing detection methods fall short in TL. We depict the feature space t-SNE results comparing the attacks in \textbf{\underline{Case-0}} and \textbf{\underline{Case-2}} in Figure \ref{fig:tsne}.
Since in TL, the model parameters have been initialized with additional knowledge obtained from pre-training, clean and poisoned samples are harder to be separated in the embedding space, thus resulting in a worse detection result compared to SL. As shown in Figure \ref{fig:tsne}, for both BadNets and the Blended attack, the clean and poisoned samples have a larger overlapping in the TL case than in SL. These results 
emphasize the importance of introducing active measures to increase separability.

On the other hand, for all the evaluated settings on the two datasets (CIFAR-10 and STL-10, Appendix \ref{sec:newres}), our method consistently achieves the best TPR, FPR, ASR, and ACC. 
Remarkably, the averaging performance on both upstream and downstream of $\AlgName$ is of magnitude better than the seven baselines. The results highlight that actively introducing different model behaviors can help a detection method to be of better robustness to the DL paradigm shift.


\renewcommand\arraystretch{2.5} 
\begin{table}[h!]
\large
\centering
\resizebox{1.0\columnwidth}{!}{\begin{tabular}{c cc|cc|cc|cc||cc||cc}
\hline
\multicolumn{1}{c|}{} & \multicolumn{4}{c|}{\textsf{\huge FT-all}}         & \multicolumn{4}{c||}{\textsf{\huge FT-last}}
& \multicolumn{2}{c}{\multirow{2}{*}{\textbf{\huge Average}}}
& \multicolumn{2}{||c}{\multirow{2}{*}{\textbf{\huge Worst-Case}}}\\ \cline{2-9} 
\multicolumn{1}{c|}{} & \multicolumn{2}{c|}{\textbf{\Large BadNets (20\%)}} & \multicolumn{2}{c|}{\textbf{\Large SAA (5\%)}} & \multicolumn{2}{c|}{\textbf{\Large Blended (20\%)}} & \multicolumn{2}{c||}{\textbf{\Large HTBA (5\%)}}  & \multicolumn{2}{c}{} & \multicolumn{2}{||c}{}
\\ \hline 

                  \multicolumn{13}{c}{
  \textbf{\huge (\textcolor{blue}{a}) Upstream Evaluation}} \\
   \hline
                 \multicolumn{1}{c|}{} & \Large TPR $\uparrow$  & \Large FPR $\downarrow$  & \Large TPR $\uparrow$  & \Large FPR $\downarrow$  & \Large TPR $\uparrow$  & \Large FPR $\downarrow$  & \Large TPR $\uparrow$  & \Large FPR $\downarrow$  & \Large TPR $\uparrow$  & \Large FPR $\downarrow$ & \Large TPR $\uparrow$  & \multicolumn{1}{|c}{\Large FPR $\downarrow$} \\ \hline

  \multicolumn{1}{c|}{\textbf{Spectral}}
          & \huge 82.3 & \huge 16.9 & \huge 39.2 & \huge 5.83 & \textcolor{minired}{\huge 11.6} & \textcolor{minired}{\huge 34.6} & \huge 53.6 & \huge 5.07 & \huge 46.7 & \huge 15.6&\huge 11.6&\multicolumn{1}{|c}{\huge 34.6}\\ \hline
          
\multicolumn{1}{c|}{\textbf{Spectre }}          & \huge 85.1 & \huge 16.2 & \huge \textcolor{minired}{53.6} & \huge 4.54 & \huge 68.5 & \huge \textcolor{minired}{20.4} & \huge 74.4 & \huge 3.98 & \huge 70.4 & \huge 11.3& \huge53.6& \multicolumn{1}{|c}{\huge20.4}\\ \hline

\multicolumn{1}{c|}{\textbf{Beatrix}}          & \huge 64.4 & \huge 19.1 & \huge 66.8 & \huge 3.96 & \huge \textcolor{minired}{13.1} & \huge \textcolor{minired}{31.4} & \huge 89.6 & \huge 3.50 & \huge 58.5 & \huge 14.5& \huge13.1& \multicolumn{1}{|c}{\huge31.4}\\ \hline

\multicolumn{1}{c|}{\textbf{AC }}               & \huge 21.6 & \huge 46.3 & \huge 57.2 & \huge 32.5 & \huge \textcolor{minired}{0.60}  & \huge \textcolor{minired}{46.9} & \huge 41.6 & \huge 34.4 & \huge 30.3 & \huge 40.0& \huge0.60& \multicolumn{1}{|c}{\huge46.9} \\ \hline

\multicolumn{1}{c|}{\textbf{ABL }}              & \huge 59.8 & \huge 22.6 & \huge \textcolor{minired}{48.4} & \huge 5.35 & \huge 49.3 & \huge \textcolor{minired}{25.2} & \huge 61.2 & \huge 4.67 & \huge 54.7 & \huge 14.5 & \huge48.4& \multicolumn{1}{|c}{\huge25.2}\\ \hline

\multicolumn{1}{c|}{\textbf{STRIP  }}           & \huge \textcolor{minired}{92.3}    & \huge 10.6 & \huge 25.6 & \huge 8.23 & \huge \textcolor{minired}{67.1}    & \huge \textcolor{minired}{16.8} & \huge 35.6 & \huge 8.70 & \huge 55.2 & \huge 11.1& \huge35.6& \multicolumn{1}{|c}{\huge16.8}\\ \hline

\multicolumn{1}{c|}{\textbf{CT   }}             & \huge 94.6 & \huge \textcolor{minired}{10.4} & \huge 78.0   & \huge 7.24 & \huge \textcolor{minired}{0.00}    & \huge 0.00    & \huge 82.4 & \huge 3.49 & \huge 63.8 & \huge 5.33& \huge0.00& \multicolumn{1}{|c}{\huge10.4} \\ \hline

\multicolumn{1}{c|}{\textbf{Ours}}              & \textcolor{minired}{\huge 98.7} & \textcolor{minired}{\huge 1.03} & \huge 95.2 & \huge 0.51 & \huge 99.2 & \huge 0.10  &  \huge 95.6 & \huge 0.34 & \textbf{\huge 97.2} & \textbf{\huge 0.50}& \textbf{\huge 95.2} & \multicolumn{1}{|c}{\textbf{\huge 1.03}}\\ \hline

\multicolumn{13}{c}{\textbf{\huge (\textcolor{blue}{b}) Downstream Evaluation}} \\
   \hline
\multicolumn{1}{c|}{} & \multicolumn{1}{c}{\Large ASR $\downarrow$} & \Large ACC $\uparrow$  & \multicolumn{1}{c}{\Large ASR $\downarrow$}  & \Large ACC $\uparrow$  &  \multicolumn{1}{c}{\Large ASR $\downarrow$}  & \Large ACC $\uparrow$  & \multicolumn{1}{c}{\Large ASR $\downarrow$}  & \Large ACC $\uparrow$  & \Large ASR $\downarrow$ & \Large ACC $\uparrow$ & \Large ASR $\downarrow$ & \multicolumn{1}{|c}{\Large ACC $\uparrow$}  \\ \hline

\multicolumn{1}{c|}{\textbf{No Def.}}    & \cellcolor[HTML]{F3CECA}\textnormal{\huge 97.5} & \huge 91.3 & \cellcolor[HTML]{F3CECA}\textcolor{minired}{\huge 98.7} & \huge 92.3 & \cellcolor[HTML]{F3CECA}\textnormal{\huge 93.9} & \textcolor{minired}{\huge 71.4}  & \cellcolor[HTML]{F3CECA}\textnormal{\huge 56.4} & \huge 72.8 
& \huge 86.6 & \huge 82.0 & \huge 98.7 & \multicolumn{1}{|c}{\huge 71.4}\\ \hline

\multicolumn{1}{c|}{\textbf{Spectral }}          & \huge \cellcolor[HTML]{F3CECA}{\textcolor{minired}{97.4}} & \huge 91.5 & \huge \cellcolor[HTML]{F3CECA}{80.2} & \huge 91.8 & \huge \cellcolor[HTML]{F3CECA}{91.4} & \huge \textcolor{minired}{68.7} & \huge \cellcolor[HTML]{D5E2F1}{16.9} & \huge 72.1  
& \huge 71.5 & \huge 81.0 & \huge97.4& \multicolumn{1}{|c}{\huge68.7}\\ \hline

\multicolumn{1}{c|}{\textbf{Spectre }}          & \huge \cellcolor[HTML]{F3CECA}{\textcolor{minired}{95.8}} & \huge 91.8 & \huge \cellcolor[HTML]{F3CECA}{75.9} & \huge 91.9 & \huge \cellcolor[HTML]{F3CECA}{92.5} & \huge \textcolor{minired}{69.8} & \huge \cellcolor[HTML]{D5E2F1}{10.9} & \huge 72.3  
& \huge 68.8 & \huge 81.5& \huge95.8& \multicolumn{1}{|c}{\huge69.8}\\ \hline

\multicolumn{1}{c|}{\textbf{Beatrix}}          	& \huge \cellcolor[HTML]{F3CECA}{\textcolor{minired}{96.0}} & \huge 91.7 & \huge \cellcolor[HTML]{F3CECA}{68.9} & \huge 92.0 & \huge \cellcolor[HTML]{F3CECA}{92.7} & \huge \textcolor{minired}{67.6} & \huge \cellcolor[HTML]{D5E2F1}{5.50} & \huge 72.6   
& \huge 65.8 & \huge 81.0 & \huge96.0& \multicolumn{1}{|c}{\huge67.6}\\ \hline

\multicolumn{1}{c|}{\textbf{AC }}               & \huge \cellcolor[HTML]{F3CECA}{\textcolor{minired}{97.4}} & \huge 86.7 & \huge \cellcolor[HTML]{F3CECA}{73.2} & \huge 88.7 & \huge \cellcolor[HTML]{F3CECA}{93.3} & \huge \textcolor{minired}{65.4} & \huge \cellcolor[HTML]{F3CECA}{21.4} & \huge 66.1  
& \huge 71.3 & \huge 76.7 & \huge97.4& \multicolumn{1}{|c}{\huge65.4}\\ \hline

\multicolumn{1}{c|}{\textbf{ABL }}               		& \huge \cellcolor[HTML]{F3CECA}{\textcolor{minired}{96.4}} & \huge 91.7 & \huge \cellcolor[HTML]{F3CECA}{80.1} & \huge 92.0 & \huge \cellcolor[HTML]{F3CECA}{93.7} & \huge \textcolor{minired}{68.3} & \huge \cellcolor[HTML]{D5E2F1}{14.2} & \huge 72.2  
& \huge 71.1 & \huge 81.1& \huge96.4& \multicolumn{1}{|c}{\huge68.3}\\ \hline

\multicolumn{1}{c|}{\textbf{Strip }}            		& \huge \cellcolor[HTML]{F3CECA}{\textcolor{minired}{94.4}} & \huge 91.8 & \huge \cellcolor[HTML]{F3CECA}{87.0} & \huge 91.9 & \huge \cellcolor[HTML]{F3CECA}{92.9} & \huge \textcolor{minired}{70.8} & \huge \cellcolor[HTML]{F3CECA}{24.3} & \huge 71.3  
& \huge 74.7 & \huge 81.5 & \huge 94.4 & \multicolumn{1}{|c}{\huge 70.8}\\ \hline

\multicolumn{1}{c|}{\textbf{CT }}               		& \cellcolor[HTML]{F3CECA}{\huge 93.2} & \huge 91.8 & \cellcolor[HTML]{D5E2F1}{\huge 18.6} & \huge 91.9 & \cellcolor[HTML]{F3CECA}{\textcolor{minired}{\huge 93.9}} & \textcolor{minired}{\huge 71.4} & \cellcolor[HTML]{D5E2F1}{\huge 8.60} & \huge 72.5  
& \huge 53.6 & \huge 81.9 & \huge 93.9 & \multicolumn{1}{|c}{\huge 71.4}\\ \hline

\multicolumn{1}{c|}{\textbf{Ours}}            &  \cellcolor[HTML]{D5E2F1}{\huge 10.2}  & \huge 92.9  &  \cellcolor[HTML]{D5E2F1}{\huge 8.40} & \huge 92.3  & \cellcolor[HTML]{D5E2F1}{\textcolor{minired}{\huge 16.2}} & \huge 74.8 & \cellcolor[HTML]{D5E2F1}{\huge 3.40} & \textcolor{minired}{\huge 72.8} & \textbf{\huge 9.55} & \textbf{\huge 83.2}& \textbf{\huge 16.2} & \multicolumn{1}{|c}{\textbf{\huge 72.8}} \\ \hline
\end{tabular}}
\caption{
(\textcolor{blue}{a}) Upstream and (\textcolor{blue}{b}) Downstream Evaluation and comparison results under \textbf{\underline{Case-2}} with CIFAR-10:
The first row denotes the TL strategy. 
The \textbf{bolded} results denote the best defense results among all defenses. We highlight the ASR below 20\% in \scalebox{0.9}{\colorbox[HTML]{D5E2F1}{\textbf{blue}}} as a success defense, the ASR above 20\% in \scalebox{0.9}{\colorbox[HTML]{F3CECA}{\textbf{red}}} as a failed defense case.
}
\label{tab:TLcifar10}
\vspace{-.5em}
\end{table}
\renewcommand\arraystretch{1}

\vspace{-1.em}
\subsection{Adaptive Attack Analysis}
\label{sec:adptiveattack}


From the above, we find $\AlgName$ is the most reliable detection method across different attacks, datasets, poison ratios, and training paradigms. Now we study adaptive attacks, where we want to understand how an attacker's knowledge about defense implementation impacts defense performance. 

\noindent
\textbf{Attacker goal \& settings.}
The attacker aims to craft poisoned samples
resulting in a low TPR while maintaining a low FPR for upstream detection
, and resulting in a high ASR while maintaining a high ACC for the downstream poisoned model.
A successful adaptive attack should achieve satisfying results based on these metrics simultaneously. 
We consider two models of attack knowledge: \emph{\underline{White-box}} attack and \emph{\underline{Gray-box}} attack. (1)
\emph{\underline{White-box}} Settings.
The attacker has full access to the details of $\AlgName$, namely, the workflow of $\AlgName$; the architecture of the detector model, 
and the architecture of the feature extractor;
the architecture of the weighting network will be used for poison concentration;
the original poisoned dataset, $D_{\text{poi}}$; and the clean base set $D_{\text{b}}$. 
Although such disclosure of the defense details is rare in practice, an investigation of this setting gives insights into the worst-case performance of $\AlgName$. (2) \emph{\underline{Gray-box}} Settings. We also consider a more realistic attack scenario where the attacker is aware of the $\AlgName$ pipeline and the respective datasets but not aware of the specific model architectures used by the defender for conducting the detection and performing downstream tasks. 
In both \emph{\underline{White-box}} attack and \emph{\underline{Gray-box}}, the attacker updates the original poisoned samples in $D_{\text{poi}}$ and then supplies the updated dataset to the defender. 
%

\begin{table*}[t!]
    \vspace{-1.5em}
    \centering
    \resizebox{0.9\textwidth}{!}{
    \begin{tabular}{cp{0.96cm}<{\centering}p{0.96cm}<{\centering}|p{0.96cm}<{\centering}p{0.96cm}<{\centering}|p{0.96cm}<{\centering}p{0.96cm}<{\centering}|p{0.96cm}<{\centering}p{0.96cm}<{\centering}|p{0.96cm}<{\centering}p{0.96cm}<{\centering}|p{0.96cm}<{\centering}p{0.96cm}<{\centering}|p{0.96cm}<{\centering}p{0.96cm}<{\centering}}
    \hline
    \multicolumn{1}{c|}{} &
      \multicolumn{8}{c|}{\cellcolor[HTML]{FFFFFF}{\color[HTML]{000000} \textit{Dirty-Label Backdoor Attacks}}} &
      \multicolumn{6}{c}{\cellcolor[HTML]{FFFFFF}{\color[HTML]{000000} \textit{Clean-Label Backdoor Attacks}}} \\ \cline{2-15} 
    \multicolumn{1}{c|}{} &
      \multicolumn{2}{c|}{\textbf{BadNets (5\%)}} &
      \multicolumn{2}{c|}{\textbf{Blended (5\%)}} &
      \multicolumn{2}{c|}{\textbf{WaNet (10\%)}} &
      \multicolumn{2}{c|}{\textbf{ISSBA (1\%)}} &
      \multicolumn{2}{c|}{\textbf{LC (1\%)}} &
      \multicolumn{2}{c|}{\textbf{SAA (1\%)}} &
      \multicolumn{2}{c}{\textbf{Narci. (0.05\%)}} \\ \hline 
    \multicolumn{15}{c}{
\multirow{2}{*}{\textbf{(\textcolor{blue}{a}) Upstream Evaluation}}} \\ \\
   \hline
   
    \multicolumn{1}{c|}{} &
      TPR $\uparrow$&
      FPR $\downarrow$&
      TPR $\uparrow$&
      FPR $\downarrow$&
      TPR $\uparrow$&
      FPR $\downarrow$&
      TPR $\uparrow$&
      FPR $\downarrow$&
      TPR $\uparrow$&
      FPR $\downarrow$&
      TPR $\uparrow$&
      FPR $\downarrow$&
      TPR $\uparrow$&
      FPR $\downarrow$\\ \hline
      
    \multicolumn{1}{c|}{\emph{\underline{White-box}}} &
      60.6 &
      1.47 &
      98.1 &
      0.49 &
      65.3 &
      5.41 &
      80.6 &
      0.03 &
      41.4 &
      37.3 &
      85.4 &
      0.14 &
      36.0 &
      17.6 \\ \hline

    \multicolumn{1}{c|}{\emph{\underline{Gray-box}}} &
      99.7 &
      0.22 &
      99.6 &
      0.18 &
      83.4 &
      4.18 &
      90.6 &
      0.08 &
      98.6 &
      0.57 &
      96.4 &
      47.1 &
      100 &
      0.03 \\ \hline
      
    \multicolumn{15}{c}{
\multirow{2}{*}{\textbf{(\textcolor{blue}{b}) Downstream Evaluation}}} \\ \\
   \hline
   
      \multicolumn{1}{c|}{} &
        ASR $\downarrow$&
        ACC $\uparrow$&
        ASR $\downarrow$&
        ACC $\uparrow$&
        ASR $\downarrow$&
        ACC $\uparrow$&
        ASR $\downarrow$&
        ACC $\uparrow$&
        ASR $\downarrow$&
        ACC $\uparrow$&
        ASR $\downarrow$&
        ACC $\uparrow$&
        ASR $\downarrow$&
        ACC $\uparrow$\\ \hline
\multicolumn{15}{c}{
\emph{\underline{White-box}} Adaptive Attack
}
\\
\hline

    \multicolumn{1}{c|}{\textbf{No Defense}} &
        \cellcolor[HTML]{F3CECA}93.1 &
        \multicolumn{1}{c|}{93.5} &
        \cellcolor[HTML]{F3CECA}83.7 &
        \multicolumn{1}{c|}{93.9} &
        \cellcolor[HTML]{F3CECA}41.2 &
        \multicolumn{1}{c|}{92.9} &
        \cellcolor[HTML]{F3CECA}84.1 &
        93.8 &
        \cellcolor[HTML]{F3CECA}95.6 &
        \multicolumn{1}{c|}{94.2} &
        \cellcolor[HTML]{F3CECA}34.2 &
        \multicolumn{1}{c|}{{93.7}} &
        \cellcolor[HTML]{F3CECA}25.3 &
        {94.7} \\ \hline

    \multicolumn{1}{c|}{\textbf{Ours}} &
        \cellcolor[HTML]{F3CECA}58.3 &
        \multicolumn{1}{c|}{94.5} &
        \cellcolor[HTML]{D5E2F1}8.41 &
        \multicolumn{1}{c|}{94.1} &
        \cellcolor[HTML]{D5E2F1}11.4 &
        \multicolumn{1}{c|}{93.3} &
        \cellcolor[HTML]{F3CECA}22.5 &
        94.4 &
        \cellcolor[HTML]{F3CECA}20.3 &
        \multicolumn{1}{c|}{93.9} &
        \cellcolor[HTML]{D5E2F1}2.35 &
        \multicolumn{1}{c|}{94.4} &
        \cellcolor[HTML]{D5E2F1}5.49 &
        94.9 \\ \hline

    \multicolumn{1}{c|}{\textbf{Ours + Unlearn}} &
        \cellcolor[HTML]{D5E2F1}3.21 &
        \multicolumn{1}{c|}{89.4} &
        \cellcolor[HTML]{D5E2F1}0.46 &
        \multicolumn{1}{c|}{91.2} &
        \cellcolor[HTML]{D5E2F1}2.45 &
        \multicolumn{1}{c|}{88.7} &
        \cellcolor[HTML]{D5E2F1}0.87 &
        93.6 &
        \cellcolor[HTML]{D5E2F1}0.53 &
        \multicolumn{1}{c|}{71.2} &
        \cellcolor[HTML]{D5E2F1}0.63 &
        \multicolumn{1}{c|}{92.3} &
        \cellcolor[HTML]{D5E2F1}1.21 &
        92.0 \\ 
\hline
\multicolumn{15}{c}{
\emph{\underline{Gray-box}} Adaptive Attack
}
\\
\hline
    \multicolumn{1}{c|}{\textbf{No Defense}} &
        \cellcolor[HTML]{F3CECA}91.3 &
        \multicolumn{1}{c|}{90.5} &
        \cellcolor[HTML]{F3CECA}88.5 &
        \multicolumn{1}{c|}{91.5} &
        \cellcolor[HTML]{F3CECA}83.6 &
        \multicolumn{1}{c|}{89.8} &
        \cellcolor[HTML]{F3CECA}26.2 &
        90.3 &
        \cellcolor[HTML]{F3CECA}64.6 &
        \multicolumn{1}{c|}{91.0} &
        \cellcolor[HTML]{D5E2F1}15.2 &
        \multicolumn{1}{c|}{{91.1}} &
        \cellcolor[HTML]{F3CECA}22.4 &
        {91.1} \\ \hline
    
    \multicolumn{1}{c|}{\textbf{Ours}} &
        \cellcolor[HTML]{D5E2F1}6.23 &
        \multicolumn{1}{c|}{90.9} &
        \cellcolor[HTML]{D5E2F1}4.35 &
        \multicolumn{1}{c|}{90.6} &
        \cellcolor[HTML]{D5E2F1}8.64 &
        \multicolumn{1}{c|}{89.0} &
        \cellcolor[HTML]{D5E2F1}1.23 &
        90.3 &
        \cellcolor[HTML]{D5E2F1}8.57 &
        \multicolumn{1}{c|}{91.1} &
        \cellcolor[HTML]{D5E2F1}1.06 &
        \multicolumn{1}{c|}{{91.1}} &
        \cellcolor[HTML]{D5E2F1}1.34 &
        {91.1} \\ \hline      

    \end{tabular}}
    \caption{(\textcolor{blue}{a}) Upstream and (\textcolor{blue}{b}) Downstream evaluation results for the adaptive attacks. We consider the same attacks from \textbf{\underline{Case-0}}, CIFAR-10, and implement the white-box adaptive attack to disguise the original poisoned samples. 
    }
    \label{tab:adaptiveup}
    \vspace{-1.5em}
    \end{table*}

\noindent
\textbf{Attack design.} 
For both \emph{\underline{White-box}} and \emph{\underline{Gray-box}} attack, we investigate optimization-based techniques to design poisoned samples to evade $\AlgName$. 
The attacker can use $D_{\text{poi}}$ and $D_{\text{b}}$ to obtain trained detector parameters, $\theta_{I}$ 
and then resolve the following optimization to obtain an additive noise for each poisoned sample $x_{\text{poi}}$ in $D_\text{poi}$ to evade the detection
\vspace{-0.8em}
\begin{equation}
\delta^* = \arg\min_{\delta} 
\mathcal{L}_{\text{max}}\left(f(x_{\text{poi}}+\delta|\theta_I)\right),
\label{eqn:adp_atk}
\vspace{-1em}
\end{equation}
where $\mathcal{L}_{\text{max}}$ is inherited from Eqn. (\ref{eqn:offset}).
Recall that $\AlgName$ optimizes $\theta$ so that poisoned samples are assigned with large loss values while clean samples are assigned with small loss values. 
The above formulation manipulates one poisoned sample, $x_{\text{poi}}$, such that the trained detector will assign low loss values to $x_{\text{poi}}+\delta^*$, which helps disguise the poison. 
To resolve the proposed adaptive attack formulation in Eqn (\ref{eqn:adp_atk}), we conduct gradient descent 100 steps for each example.  
%
Visual examples of the adaptive attack manipulated poisons for the attacks considered in $\textbf{\underline{Case-0}}$ are depicted in Figure \ref{fig:adaptattack}, Appendix \ref{sec:newres}.
After the update of $D_{\text{poi}}$, we obtain new model parameters (i.e., feature extractor $\tilde{\theta}_{\text{poi}}^*$, detector $\tilde{\theta}_I$, and weighting network $\tilde{M}$) on the updated $D_{\text{poi}}$ and evaluate the attack performance following the aforementioned attack settings. 
For the \emph{\underline{White-box}} attack setting, we evaluate the downstream with the same model structures as
used by the attacker for synthesizing $\delta^*$. For the \emph{\underline{Gray-box}} attack setting, we use different model structures.

\noindent
\textbf{Results and insights.}
The results of $\AlgName$ against the adaptive attacks are summarized in Table \ref{tab:adaptiveup}. 
From the upstream evaluation, for \emph{\underline{White-box}} attack, we find the adaptive attack's effect varies from trigger to trigger. The performance of the \emph{\underline{White-box}} attack on disguising Blended triggers is limited, while on BadNets, LC, and Narci., the TPR is largely decreased. 
Interestingly, the model mismatch introduced in the \emph{\underline{Gray-box}} largely impacts the attack efficacy and $\AlgName$ is able to maintain high defense performance across all the \emph{\underline{Gray-box}} attacks.
While moving on to the downstream evaluation, we find both \emph{\underline{White-box}} and \emph{\underline{Gray-box}} adaptive attack introduced additional noise that impedes some of the backdoor triggers from taking effect, i.e., lower ASR at the end, even without any additional defensive measure. 
For \emph{\underline{White-box}} attack, we find only the adaptive BadNets attack can achieve an ASR greater than 50\% after the model converges over the subset removing the detected samples using $\AlgName$. By following the standard procedure in many detection-based defenses\cite{li2021anti,qi2022fight}, we use the detected samples to provide revered gradients for the downstream model (e.g., minimize negative CE loss) or known as Unlearning, denoted by ``Ours+Unlearn''.
We find this simple adaptation of $\AlgName$ can successfully diminish the effect of all the evaluated \emph{\underline{White-box}} adaptive attacks. 
On the other hand, the $\AlgName$ on the \emph{\underline{Gray-box}} adaptive attacks with detector model mismatch (attacker uses ResNet-18 to obtain $\theta_I$, defender uses VGG-16 to obtain $\tilde{\theta}_I$) are almost the same on the vanilla attacks without adaptation.

To conclude, the above study shows that $\AlgName$ is robust to the evaluated \emph{\underline{White-box}} attack with the standard unlearning procedure using the detected samples and robust to the evaluated \emph{\underline{Gray-box}} attack.
The results highlight that disclosing the knowledge of our defense workflow and models can expose $\AlgName$ to the risk of adaptive attacks. Not releasing the model architecture can mitigate the risk of adaptive attacks to a large extent. Also, using the detected samples for unlearning can be a simple yet effective post-processing method that can be used in tandem with our detection to safeguard ML applications against adaptive attacks to our defense. One thing worth highlighting is that the unlearning process requires the detection method to obtain a better 
precision upstream.
Otherwise, if the FPR of the upstream is high (more clean samples are wrongly flagged), the downstream unlearning would result in an unfavorable impact on the ACC (e.g. the results on the \emph{\underline{White-box}} LC results).


\vspace{-.5em}
\section{Conclusion}
\vspace{-.5em}
This work is motivated by the glaring gap between the focused evaluation of the end-to-end SL settings in prior backdoor detection literature and the fast adaption of other more data- and computation-efficient learning paradigms, including SSL adaptation and TL. We find that existing detection methods cannot be applied or suffer limited performance for SSL and TL; even for the widely studied end-to-end SL setting, there is still large room to improve detection in terms of their robustness to variations in poison ratio. This work proposes a novel idea for actively enforcing different model behaviors on clean and poisoned samples through a two-level nested offset loop. Our approach provides the first backdoor defense that operates across different learning paradigms, different attack techniques, and poison ratios.

Our work opens up many directions for future work. (1) \emph{Theoretical Understanding of Offset}: Despite the empirical success, an in-depth understanding of convergence behaviors and sample complexity of $\AlgName$ is still lacking. In addition, we have shown multiple offset objectives, but how to explain why a loss design is better than the other is still an open question. (2) \emph{Alternative Offset Goal Designs}: Our work provides a general algorithmic framework for active backdoor data detection by optimizing opposite goals. Are there other optimization objectives beyond what we proposed in this paper that can lead to better detection performance? (3) \emph{Extension to Broader Data Types}: Evaluating $\AlgName$ on domains beyond images and texts is of practical importance.

\vspace{-1em}
\section*{Acknowledgement}
\vspace{-.5em}
RJ and the ReDS lab appreciate the support of the Amazon - Virginia Tech Initiative for Efficient and Robust Machine Learning and the Cisco Award. YZ is supported by the Amazon Fellowship. XL gratefully acknowledges the support of National Science Foundation Award No. CNS-1929300.
\vspace{-1em}



\bibliographystyle{IEEEtran}
\bibliography{bibtex}


\vspace{-1em}
\section*{Appendix}

\subsection{Detailed Defense Settings}
\label{sec:defense_set}
\vspace{-.5em}
In the evaluation section, we provide a thorough comparison of existing backdoor detection techniques. These methods can be classified into several categories, including Spectral\cite{tran2018spectral}, 
Spectre\cite{hayase2021spectre}, 
and Beatrix\cite{ma2022beatrix}, 
which utilize analysis of activation patterns; AC\cite{chen2018detecting}, 
which leverages clustering of feature information; ABL\cite{li2021anti}, 
which detects the lowest loss from poisoned datasets; Strip\cite{gao2019strip}, 
which focuses on logits of sample outputs; and CT\cite{qi2022fight},
which employs confusion training in end-to-end supervised learning settings. 

Note that the above baseline defenses were only evaluated under the settings of end-to-end SL (\textbf{\underline{Case-0}}) in their original papers. They can also be directly generalized to \textbf{\underline{Case-2}}. We will incorporate the above seven baseline defenses in \textbf{\underline{Case-0}} and \textbf{\underline{Case-2}} with the suggested hyperparameters proposed in these original works for comparison.
As for \textbf{\underline{Case-1}}, some of the methods are not applicable, whereas others can be adapted to operate without label information.
In particular, Strip \cite{gao2019strip} and CT \cite{qi2022fight} are label-information-dependent methods, which are excluded from evaluation in \textbf{\underline{Case-1}}.
The vanilla design of Spectral \cite{tran2018spectral} and Spectre \cite{hayase2021spectre} used a feature extractor trained with label information. In our \textbf{\underline{Case-1}} experiment, we replace the feature extractor trained with labels with one trained using the SSL paradigm. The original implementation processes samples class-wisely for the Beatrix \cite{ma2022beatrix} and AC \cite{chen2018detecting}. However, since there is no label information in \textbf{\underline{Case-1}}, we process all training samples together. For ABL \cite{li2021anti}, we replace the original implementation's Cross-Entropy loss with the respective training loss function used in the respective SSL algorithm (e.g., the InfoNCE loss for the MoCo V3 \cite{chen2021empirical}).

\vspace{-1em}
\subsection{Detailed Attack Settings}
\label{sec:attack_set}
\vspace{-.5em}


In this work, we examine several representative attacks for each category of attack design. For \textbf{\underline{Case-0}}, which is the end-to-end supervised learning setting mentioned in Section \ref{sec:case0exp}, we thoroughly investigate existing Dirty-label attacks and Clean-label backdoor attacks. Dirty-label attacks create a backdoor by altering the label of the poisoned samples to the target class. We selected some representative attacks for experiments. For example, BadNets\cite{gu2017badnets} and Blended\cite{chen2017targeted} are used as triggers by simply superimposing special patterns; there are also affine transformations that are difficult to find on pictures, such as WaNet\cite{nguyen2021wanet};
as well as training an encoder to create distinct backdoor trigger for each sample like ISSBA\cite{li2020invisible}. 
On the other hand, Clean-label backdoor attacks maintain the original label of the poisoned samples. Examples include LCcite{turner2019label}, 
which makes models learn simple triggers by patching adversarial noise on the remaining part of sample; SAA\cite{souri2021sleeper},
which produces effects through model feature collisions; and the state-of-the-art attack Narcissus\cite{zeng2022narcissus}, 
which obtains the backdoor trigger by optimizing the distribution within the class and the connection of the target label. For these three Clean-label backdoor attacks, we set $l_{\infty}=16/255$ to ensure the consistency of the attack. 
For \textbf{\underline{Case-1}}, we consider the backdoor attack in the SSL setting (detailed in Section \ref{sec:sslexp}). Since the training does not require labels and always contains strong augmentations, traditional attacks against SSL are not effective. However, with the development of this training paradigm, attacks against it have started to emerge. There are attacks by superimposing specific design patterns\cite{carlini2021poisoning, saha2022backdoor}
and attacks by adding specific frequency noise to the YCbCr color space\cite{li2022demystifying}. The C-brd and C-Squ adopt a fixed in-class poison ratio w.r.t. only the samples from the targeted category (50\% in-class), following \cite{saha2022backdoor}. CTRL adopts a fixed poison ratio w.r.t. the whole dataset (1\% of all the samples), following \cite{li2022demystifying}. 
For \textbf{\underline{Case-2}}, we investigate the attacks in the context of transfer learning, as described in section \ref{sec:finetuneexp}. Our evaluation revealed that adding backdoor attack samples to the fine-tuned dataset leads to a successful attack. Basic backdoor attacks, such as BadNets and Blended, can easily be generalized and result in an effective attack. Furthermore, attacks based on the collision of the model's feature space, such as SSA
or HTBA\cite{saha2020hidden}
can also work in this scenario. 
All the attacks use the default settings in the original paper to ensure consistency with the original work.


 \vspace{-1em}
\subsection{Additional results}
\label{sec:newres}
\vspace{-.5em}

\begin{table*}[t!]
\centering
\vspace{-1.5em}
\resizebox{0.8\textwidth}{!}{
\begin{tabular}{
>{\columncolor[HTML]{FFFFFF}}c 
>{\columncolor[HTML]{FFFFFF}}c 
>{\columncolor[HTML]{FFFFFF}}c 
>{\columncolor[HTML]{FFFFFF}}c ||
>{\columncolor[HTML]{FFFFFF}}c 
>{\columncolor[HTML]{FFFFFF}}c 
>{\columncolor[HTML]{FFFFFF}}c 
>{\columncolor[HTML]{FFFFFF}}c |
>{\columncolor[HTML]{FFFFFF}}c 
>{\columncolor[HTML]{FFFFFF}}c 
>{\columncolor[HTML]{FFFFFF}}c 
>{\columncolor[HTML]{FFFFFF}}c |
>{\columncolor[HTML]{FFFFFF}}c 
>{\columncolor[HTML]{FFFFFF}}c 
>{\columncolor[HTML]{FFFFFF}}c 
>{\columncolor[HTML]{FFFFFF}}c }
\hline

 \multicolumn{1}{c|}{} &
  \multicolumn{3}{c||}{\cellcolor[HTML]{FFFFFF}{\color[HTML]{000000} \textbf{No Attack}}} &
  \multicolumn{4}{c|}{\cellcolor[HTML]{FFFFFF}{\color[HTML]{000000} \textbf{C-brd (5\%)}}} &
  \multicolumn{4}{c|}{\cellcolor[HTML]{FFFFFF}{\color[HTML]{000000} \textbf{C-Squ (5\%)}}} &
  \multicolumn{4}{c}{\cellcolor[HTML]{FFFFFF}{\color[HTML]{000000} \textbf{CTRL (1\%)}}} \\ 
     \hhline{~|-|-|-|-|-|-|-|-|-|-|-|-|-|-|-|}
  \multicolumn{1}{c|}{} &
  {\color[HTML]{000000} ASR} $\downarrow$&
  {\color[HTML]{000000} ASR*} $\downarrow$&
  {\color[HTML]{000000} ACC} $\uparrow$&
  {\color[HTML]{000000} ASR*$_{0}$} &
  \cellcolor[HTML]{FFFFFF}{{\color[HTML]{000000} ASR*}} $\downarrow$&
  {\color[HTML]{000000} ACC$_{0}$} &
  {\color[HTML]{000000} ACC} $\uparrow$&
  {\color[HTML]{000000} ASR*$_{0}$} &
  \cellcolor[HTML]{FFFFFF}{{\color[HTML]{000000} ASR*}} $\downarrow$&
  {\color[HTML]{000000} ACC$_{0}$} &
  {\color[HTML]{000000} ACC} $\uparrow$&
  {\color[HTML]{000000} ASR$_{0}$} &
  \cellcolor[HTML]{FFFFFF}{{\color[HTML]{000000} ASR}} $\downarrow$&
  {\color[HTML]{000000} ACC$_{0}$} &
  {\color[HTML]{000000} ACC} $\uparrow$\\
  \hline
\multicolumn{1}{c|}{\textbf{SimCLR}} &
  {\color[HTML]{000000} 0.34} &
  {\color[HTML]{000000} 6} &
  {\color[HTML]{000000} 67.2} &
  {\color[HTML]{000000} 168} &
  {\color[HTML]{000000} 8} &
  {\color[HTML]{000000} 65.9} &
  {\color[HTML]{000000} 66.9} &
  {\color[HTML]{000000} 141} &
  {\color[HTML]{000000} 12} &
  {\color[HTML]{000000} 65.6} &
  {\color[HTML]{000000} 66.8} &
  {\color[HTML]{000000} 25.0} &
  {\color[HTML]{000000} 1.36} &
  {\color[HTML]{000000} 66.1} &
  {\color[HTML]{000000} 66.8} \\ \hline
\multicolumn{1}{c|}{\textbf{MoCo V3}} &
  {\color[HTML]{000000} 0.32} &
  {\color[HTML]{000000} 6} &
  {\color[HTML]{000000} 68.4} &
  {\color[HTML]{000000} 67} &
  {\color[HTML]{000000} 8} &
  {\color[HTML]{000000} 68.0} &
  {\color[HTML]{000000} 68.2} &
  {\color[HTML]{000000} 119} &
  {\color[HTML]{000000} 12} &
  {\color[HTML]{000000} 67.8} &
  {\color[HTML]{000000} 68.2} &
  {\color[HTML]{000000} 23.6} &
  {\color[HTML]{000000} 2.12} &
  {\color[HTML]{000000} 68.2} &
  {\color[HTML]{000000} 68.3} \\ \hline
\multicolumn{1}{c|}{\textbf{BYOL}} &
  {\color[HTML]{000000} 0.36} &
  {\color[HTML]{000000} 7} &
  {\color[HTML]{000000} 67.1} &
  {\color[HTML]{000000} 290} &
  {\color[HTML]{000000} 11} &
  {\color[HTML]{000000} 66.6} &
  {\color[HTML]{000000} 67.1} &
  {\color[HTML]{000000} 263} &
  {\color[HTML]{000000} 19} &
  {\color[HTML]{000000} 66.8} &
  {\color[HTML]{000000} 66.9} &
  {\color[HTML]{000000} 40.9} &
  {\color[HTML]{000000} 1.44} &
  {\color[HTML]{000000} 66.5} &
  {\color[HTML]{000000} 66.8} \\ \hline
\multicolumn{1}{c|}{ \textbf{MAE}} &
  {\color[HTML]{000000} 0.28} &
  {\color[HTML]{000000} 5} &
  {\color[HTML]{000000} 70.2} &
  {\color[HTML]{000000} 28} &
  {\color[HTML]{000000} 6} &
  {\color[HTML]{000000} 68.9} &
  {\color[HTML]{000000} 70.1} &
  {\color[HTML]{000000} 68} &
  {\color[HTML]{000000} 8} &
  {\color[HTML]{000000} 69.1} &
  {\color[HTML]{000000} 69.9} &
  {\color[HTML]{000000} 30.7} &
  {\color[HTML]{000000} 1.66} &
  {\color[HTML]{000000} 68.7} &
  {\color[HTML]{000000} 69.3} \\ \hline
\end{tabular}}
\vspace{-1.em}
\caption{
Downstream evaluation results of our method under \textbf{\underline{Case-1}} in ImageNet-100.
}
\label{tab:SSLimagenetdown}
\vspace{-1.em}
\end{table*}

\begin{figure*}[t!]
  \centering
     \hspace{-1.5em}
  \includegraphics[width=0.6\linewidth]{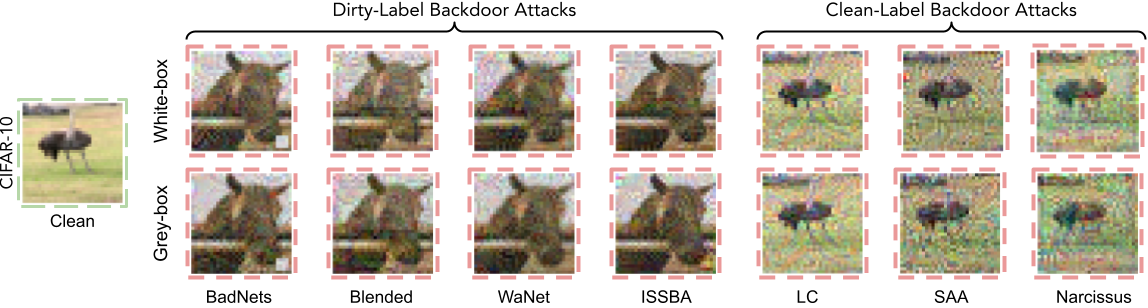}
  \vspace{-1em}
  \caption{
Visual examples of the backdoor poisoned samples disguised by adaptive attacks (\textbf{\underline{Case-0}}, CIFAR-10).
  }
  \label{fig:adaptattack}
       \vspace{-1.5em}
\end{figure*}

In addition to the results presented in the main text, we also evaluate the performance of the baseline defenses in different attack settings and dataset settings. 


\noindent
\textbf{Additional Results with Multiple Attacks.}
For \textbf{\underline{Case-0}}, we test the scenario where multiple backdoor attacks appear simultaneously in a training set. 
We deploy 4 different dirty label attacks that have appeared in the main text into 4 different classes of the CIFAR-10 dataset, and the poison ratio is consistent with the main text. At the same time, the ASR of all attacks is above 90\% to ensure the effectiveness of the attack.
The results are listed in Table \ref{tab:multipoi}. 
When multiple attacks are present, all the baseline defense methods except CT can maintain a reliable detection, as at least one set of poisoned samples ends up with a TPR lower than 50\%.
Our method achieves the highest average TPR among all defenses and demonstrates a better and more consistent detection performance with all the TPR above 85\% under this setting.


\noindent
\textbf{Additional Results with SSL.}
For  \textbf{\underline{Case-1}}, we evaluate the results on the ImageNet-100 dataset. 
ImageNet-100 is a subset of ImageNet-1K, consisting of 100 randomly selected classes (about 128,000 samples), which is currently the most popular benchmark dataset for self-supervised learning. All images are resized to 224x224 pixels to fit the model input. Here we use self-supervised learning methods consistent with those in Section 4.3, including the contrastive learning method SimCLR, MoCO V3, BYOL, and the masked-model training method MAE. Here all backbone models are ViT-Small/16 to obtain a satisfactory ACC.
The upstream and downstream results can be found in Table \ref{tab:SSLimagenet10}, and Table \ref{tab:SSLimagenetdown}, respectively. 
As the dataset becomes more complex compared to CIFAR-10, detection also becomes more difficult. Nevertheless, our method provides a TPR greater than 88\% in all cases. All FPRs are below 0.5\%, providing as clean samples as possible for subsequent downstream tasks and minimizing the impact on ACC. In the downstream task, our method succeeded in reducing the ASR with no significant improvement over the baseline without poison, indicating that our method was successful in removing the poison. At the same time, thanks to the extremely low FPR, the ACC of the model has seen a certain increase compared to the poisoned model.

\noindent
\textbf{Additional Results with TL.}
Finally, in \textbf{\underline{Case-2}}, we present the upstream and downstream results of STL-10 in Table \ref{tab:TLstl10}, where all images were scaled to 224x224 pixels to align with the ImageNet-1K\cite{deng2009imagenet} pre-trained ViT-Tiny/16\cite{dosovitskiy2020image} model. 
Our method consistently achieves a TPR of over 90\%, while keeping the FPR below 0.6\%. Compared to other defense methods, our method achieves the best average TPR and FPR. In the downstream tasks, which benefited from the high TPR and low FPR, our method successfully keeps all ASRs below 20\%, ensuring attacks will not effectively occur. Our method obtains the highest average value for ACC as well as ASR.

\noindent
\textbf{Visual Results of Adaptive Attacks.}
Figure \ref{fig:adaptattack} depicts the visual results of the adaptive attacks discussed in Section \ref{sec:adptiveattack}.

\begin{table}[h!]
\centering
\resizebox{0.65\columnwidth}{!}{
\begin{tabular}{c|cc||cc}
\hline
     & \multicolumn{2}{c||}{Upstream Evaluation} & \multicolumn{2}{c}{Downstream Evaluation} \\ \hline
     & \multicolumn{1}{c|}{TPR $\uparrow$}      & FPR $\downarrow$      & \multicolumn{1}{c|}{ASR $\downarrow$}       & ACC $\uparrow$      \\ \hline

AC   & \multicolumn{1}{c|}{79.8}     & 18.1     & \multicolumn{1}{c|}{68.4}      & 90.3     \\ \hline
Ours & \multicolumn{1}{c|}{100}      & 3.77     & \multicolumn{1}{c|}{10.3}      & 91.6     \\ \hline
\end{tabular}
}
\vspace{-1em}
\caption{
Textual backdoor detection, BadNets, SST-2 dataset.
}
\vspace{-1.em}
\end{table}

\noindent
\textbf{Additional Results on Other Modality.}
We provide additional results on exploring the applicability of the $\AlgName$ on detecting backdoor samples in the Natural Language Processing domain. We implemented the BadNets attack\footnote{\url{https://github.com/thunlp/OpenBackdoor}} on the SST-2 dataset with BERT\cite{devlin2018bert} as the target model.
We set the poisoning rate to be 10\%, with the trigger as "cf mn bb tq." 
We observe that $\AlgName$ can achieve good detection results.
Compared to the AC evaluated under the same settings, we find our method provides more effective detection results. One possible explanation for the AC's limited effectiveness is that the BERT model relies on pre-trained features, which limits the separability based on feature space clustering.  


\noindent
\textbf{Computation Overhead.}
Table \ref{tab:overhead} compares the computation overhead of $\AlgName$ and other baseline methods in \textbf{\underline{Case-0}}. 
\begin{table}[!h]
\resizebox{\columnwidth}{!}{
\centering
\begin{tabular}{c|c|c|c|c|c|c|c|c} 
\hline
         & Spectral & Spectre   & Beatrix  & AC       & ABL & Strip      & CT   & Ours         \\ 
\hline
CIFAR-10 & 1800+63  & 1800+137 & 1800+782 & 1800+123 & 847 & 1800+374 & 6300 & 1800+1800  \\
\hline
\end{tabular}}
\vspace{-1.em}
\caption{
Computational overhead (GTX 2080 Ti GPU seconds) under (\textbf{\underline{Case-0}}). Defense methods rely on a pre-trained poisoned model incur additional 1800s for training.}
\label{tab:overhead}
\vspace{-1em}
\end{table}

\begin{table}[t!]
\centering
\vspace{-1.5em}
\resizebox{\columnwidth}{!}{
\begin{tabular}{c|c|c|c|c|c||c||c}
\hline
\multirow{2}{*}{} & \multirow{2}{*}{\textbf{FPR}} & \textbf{\begin{tabular}[c]{@{}c@{}}BadNets (5\%)\\ Class 0\end{tabular}} & \textbf{\begin{tabular}[c]{@{}c@{}}WaNet (10\%)\\ Class 4\end{tabular}} & \textbf{\begin{tabular}[c]{@{}c@{}}ISSBA (1\%)\\ Class 6\end{tabular}} & \textbf{\begin{tabular}[c]{@{}c@{}}Blended (5\%)\\ Class 9\end{tabular}} & \textbf{Average} & \textbf{Worst-Case}\\ \cline{3-8} 
                  &                               & \textbf{TPR}                                                             & \textbf{TPR}                                                            & \textbf{TPR}                                                          & \textbf{TPR}       & \textbf{TPR} & \textbf{TPR}                                                      \\ \hline
\textbf{Spectral} & 27.2                          & 88.6                                                                     & \textcolor{minired}{0.00}                                                                    & 98.4                                                                  & 92.6 & 69.9   & 0.00                                                                 \\ \hline
\textbf{Spectre}  & 26.5                          & 94.6                                                                     & \textcolor{minired}{0.16}                                                                    & 84.2                                                                  & 98.9   & 69.5      & 0.16                                                            \\ \hline
\textbf{Beatrix}  & 2.61                          & 92.8                                                                     & 50.9                                                                    & \textcolor{minired}{42.6}                                                                  & 75.9     & 65.5     & 42.6                                                           \\ \hline
\textbf{AC}       & 42.4                          & 58.9                                                                     & 79.2                                                                    & \textcolor{minired}{4.60}                                                                   & 53.3     & 49.0    & 4.60                                                            \\ \hline
\textbf{ABL}      & 33.7                          & 89.9                                                                     & 0.64                                                                    & \textcolor{minired}{0.00}                                                                  & 6.72   & 24.3   & 0.00                                                               \\ \hline
\textbf{Strip}    & 11.2                          & \textcolor{minired}{83.6}                                                                     & 6.82                                                                    & 45.4                                                                  & \textcolor{minired}{51.7}     & 46.9   & 6.82                                                             \\ \hline
\textbf{CT}       & 1.22                          & 99.6                                                                     & 96.5                                                                   & \textcolor{minired}{81.8}                                                                  & 96.8                     & 93.7      & 81.8                                          \\ \hline
\textbf{Ours}     & 0.36                         & 99.7                                                                     & \textcolor{minired}{86.8}                                                                    & 94.2                                                                  & 100              & \textbf{95.2}                                                 & \textbf{86.8}       \\ \hline
\end{tabular}}
\vspace{-1em}
\caption{
Defense results on multi-trigger-multi-target attack under \textbf{\underline{Case-0}}, FPR refers to the overall FPR in the training dataset. 
The \textbf{bolded} results denote the best defense results among all the evaluated defenses w.r.t. each attack. 
}
\label{tab:multipoi}
\vspace{-.5em}
\end{table}

\begin{table}[t!]
\centering
\resizebox{0.65\columnwidth}{!}{
\begin{tabular}{ccc|cc|cc}
\hline
       \multicolumn{1}{c|}{} &
          \multicolumn{2}{c|}{\textbf{C-brd (0.5\%)}} &
  \multicolumn{2}{c|}{\textbf{C-Squ (0.5\%)}} &
  \multicolumn{2}{c}{\textbf{ CTRL (1\%)}} \\ 
     \hhline{~|-|-|-|-|-|-|}
  \multicolumn{1}{c|}{} &
  TPR $\uparrow$&
  FPR $\downarrow$&
  TPR $\uparrow$&
  FPR $\downarrow$&
  TPR $\uparrow$&
  FPR $\downarrow$\\
  \hline
\multicolumn{1}{c|}{\textbf{SimCLR}} &
  95.7 &
  0.17 &
  \textcolor{black}{92.9} &
  0.20 &
  \textcolor{black}{92.9} &
  \textcolor{black}{0.34} \\ \hline
\multicolumn{1}{c|}{\textbf{MoCo V3}} &
  92.5 &
  0.21 &
  \textcolor{black}{90.3} &
  \textcolor{black}{0.23} &
  91.4 &
  0.10 \\ \hline
\multicolumn{1}{c|}{\textbf{BYOL}} &
  90.5 &
  0.15 &
  \textcolor{black}{88.6} &
  0.18 &
  97.4 &
  \textcolor{black}{0.21} \\ \hline
\multicolumn{1}{c|}{ \textbf{MAE}} &
  97.8 &
  0.17 &
  \textcolor{black}{94.2} &
  \textcolor{black}{0.27} &
  98.8 &
  0.17 \\ \hline
\end{tabular}}
\vspace{-1em}
\caption{
Further upstream evaluation of our method under \textbf{\underline{Case-1}} with four training algorithms under ImageNet. 
}
\label{tab:SSLimagenet10}
\vspace{-.5em}
\end{table}

\begin{table}[t!]
\centering
\resizebox{\columnwidth}{!}{\begin{tabular}{ccc|cc|cc|cc||cc||cc}
\hline
\multicolumn{1}{c|}{} & \multicolumn{4}{c|}{\textsf{FT-all}}         & \multicolumn{4}{c||}{\textsf{FT-last}}
& \multicolumn{2}{c}{\multirow{2}{*}{\textbf{Average}}} & \multicolumn{2}{||c}{\multirow{2}{*}{\textbf{Worst-Case}}} \\ \cline{2-9} 
\multicolumn{1}{c|}{} & \multicolumn{2}{c|}{\textbf{BadNets (20\%)}} & \multicolumn{2}{c|}{\textbf{SAA (5\%)}} & \multicolumn{2}{c|}{\textbf{Blended (20\%)}} & \multicolumn{2}{c||}{\textbf{HTBA (5\%)}}  & \multicolumn{2}{c}{}  & \multicolumn{2}{||c}{}
\\ \hline
\multicolumn{13}{c}{
\multirow{2}{*}{\textbf{(\textcolor{blue}{b}) Upstream Evaluation}}} \\ \\
   \hline
                 \multicolumn{1}{c|}{} & TPR $\uparrow$  & FPR $\downarrow$  & TPR $\uparrow$  & FPR $\downarrow$  & TPR $\uparrow$  & FPR $\downarrow$  & TPR $\uparrow$  & FPR $\downarrow$  & TPR $\uparrow$  & FPR $\downarrow$ & TPR $\uparrow$  & \multicolumn{1}{|c}{FPR $\downarrow$} \\ \hline
                 
\multicolumn{1}{c|}{\textbf{Spectral}} & 54.4 & \multicolumn{1}{c|}{23.9}  & \textcolor{minired}{11.2} & 7.31  & 16.6 & \multicolumn{1}{c|}{\textcolor{minired}{33.4}} & 25.6 & 4.97 & 27.0 & 17.4& 11.2 & \multicolumn{1}{|c}{33.4}\\ \hline

\multicolumn{1}{c|}{\textbf{Spectre}}  & 76.8 & \multicolumn{1}{c|}{18.3}  & 17.2 & 6.99  & \textcolor{minired}{16.0}   & \multicolumn{1}{c|}{\textcolor{minired}{33.5}}  & 46.4 & 3.87 & 39.1 & 15.7& 16.0 & \multicolumn{1}{|c}{33.5}\\ \hline

\multicolumn{1}{c|}{\textbf{Beatrix}} & 86.9 & \multicolumn{1}{c|}{3.55}  & 74.8 & 12.1  & \textcolor{minired}{56.7} & \multicolumn{1}{c|}{11.7}  & 89.2 & \textcolor{minired}{13.5} & 76.9 & 10.2 & 56.7 & \multicolumn{1}{|c}{13.5}\\ \hline

\multicolumn{1}{c|}{\textbf{AC}}      & 34.6 & \multicolumn{1}{c|}{46.2} & 18.0   & 14.5 & 9.80  & \multicolumn{1}{c|}{\textcolor{minired}{60.3}} & \textcolor{minired}{8.40}  & 13.3 & 17.7 & 33.6 & 8.40 & \multicolumn{1}{|c}{60.3}\\ \hline

\multicolumn{1}{c|}{\textbf{ABL}}      & 81.2 & \multicolumn{1}{c|}{17.2}  & \textcolor{minired}{57.2} & 4.88  & 75.3 & \multicolumn{1}{c|}{\textcolor{minired}{18.7}}  & 75.6 & 3.91 &72.3 &11.2 & 57.2 & \multicolumn{1}{|c}{18.7} \\ \hline

\multicolumn{1}{c|}{\textbf{Strip}}   & 83.2  & \multicolumn{1}{c|}{11.2}  & \textcolor{minired}{0.00}    & 20.7 & \textcolor{minired}{52.9}    & \multicolumn{1}{c|}{\textcolor{minired}{16.3}} & 71.2 & 17.6 &51.8 &16.5 & 0.00 & \multicolumn{1}{|c}{20.7}\\ \hline

\multicolumn{1}{c|}{\textbf{CT}}     & 98.3 & \multicolumn{1}{c|}{\textcolor{minired}{10.5}}  & \textcolor{minired}{82.4} & 3.98  & 98.7 & \multicolumn{1}{c|}{6.58}  & 96.4 & 2.57 &94.0 & 5.91& 82.4 & \multicolumn{1}{|c}{10.5}\\ \hline

\multicolumn{1}{c|}{\textbf{Ours}}     & 97.7 & \multicolumn{1}{c|}{\textcolor{minired}{0.53}} & \textcolor{minired}{90.8} & 0.34 & 99.6 & \multicolumn{1}{c|}{0.18} & 99.2 & 0.19 &\textbf{96.8} &\textbf{0.31} & \textbf{90.8} & \multicolumn{1}{|c}{\textbf{0.53}}\\ \hline

\multicolumn{13}{c}{
\multirow{2}{*}{\textbf{(\textcolor{blue}{b}) Downstream Evaluation}}} \\ \\
   \hline
\multicolumn{1}{c|}{} & \multicolumn{1}{c|}{ASR $\downarrow$} & ACC $\uparrow$  & \multicolumn{1}{c|}{ASR $\downarrow$}  & ACC $\uparrow$  & \multicolumn{1}{c|}{ASR $\downarrow$}  & ACC $\uparrow$  & \multicolumn{1}{c}{ASR $\downarrow$}  & ACC $\uparrow$  & ASR $\downarrow$ & ACC $\uparrow$ & ASR $\downarrow$ & \multicolumn{1}{|c}{ACC $\uparrow$}  \\ \hline
\multicolumn{1}{c|}{\textbf{No Def.}}          						& \cellcolor[HTML]{F3CECA}{99.6}  & \textcolor{minired}{97.9}  & \cellcolor[HTML]{F3CECA}{93.6} & 98.6 & \cellcolor[HTML]{F3CECA}{\textcolor{minired}{97.7}} & 98.5  & \cellcolor[HTML]{F3CECA}{68.1} & 98.5 &89.8 &98.4 & 99.6 & \multicolumn{1}{|c}{97.9}\\ \hline
\multicolumn{1}{c|}{\textbf{Spectral}}          & \cellcolor[HTML]{F3CECA}{\textcolor{minired}{99.5}}  & \textcolor{minired}{97.6}  & \cellcolor[HTML]{F3CECA}{91.4} & 98.1 & \cellcolor[HTML]{F3CECA}{97.6} & 98.4  & \cellcolor[HTML]{F3CECA}{49.6} & 98.5 &84.5 &98.2 & 99.5 & \multicolumn{1}{|c}{97.6}\\ \hline
\multicolumn{1}{c|}{\textbf{Spectre}}          & \cellcolor[HTML]{F3CECA}{\textcolor{minired}{99.3}}  & \textcolor{minired}{98.0}  & \cellcolor[HTML]{F3CECA}{86.3} & 98.1 & \cellcolor[HTML]{F3CECA}{97.6} & 98.4  & \cellcolor[HTML]{F3CECA}{31.3} & 98.5 &78.6 &98.3 & 99.3 & \multicolumn{1}{|c}{98.0}\\ \hline
\multicolumn{1}{c|}{\textbf{Beatrix}}          	& \cellcolor[HTML]{F3CECA}{\textcolor{minired}{99.4}}  & 98.0  & \cellcolor[HTML]{F3CECA}{36.2} & \textcolor{minired}{97.9} & \cellcolor[HTML]{F3CECA}{95.2} & 98.6  & \cellcolor[HTML]{D5E2F1}{8.93} & 98.5 &59.9 &98.3 & 99.4 & \multicolumn{1}{|c}{97.9}\\ \hline
\multicolumn{1}{c|}{\textbf{AC}}                & \cellcolor[HTML]{F3CECA}{\textcolor{minired}{99.6}}  & \textcolor{minired}{97.3}  & \cellcolor[HTML]{F3CECA}{85.6} & 98.0 & \cellcolor[HTML]{F3CECA}{98.4} & 98.2  & \cellcolor[HTML]{F3CECA}{56.3} & 98.5 &85.0 &98.0 & 99.6 & \multicolumn{1}{|c}{97.3}\\ \hline
\multicolumn{1}{c|}{\textbf{ABL}}               		& \cellcolor[HTML]{F3CECA}{\textcolor{minired}{99.6}}  & \textcolor{minired}{97.1}  & \cellcolor[HTML]{F3CECA}{59.6} & 98.2 & \cellcolor[HTML]{F3CECA}{94.2} & 98.5  & \cellcolor[HTML]{D5E2F1}{14.3} & 98.5 &66.9 &98.1 & 99.6 & \multicolumn{1}{|c}{97.1}\\ \hline
\multicolumn{1}{c|}{\textbf{Strip}}            		& \cellcolor[HTML]{F3CECA}{\textcolor{minired}{99.5}}  & \textcolor{minired}{97.7}  & \cellcolor[HTML]{F3CECA}{94.0} & 97.8 & \cellcolor[HTML]{F3CECA}{98.4} & 98.4  & \cellcolor[HTML]{D5E2F1}{16.8} & 98.4 &77.2 &98.1& 99.5 & \multicolumn{1}{|c}{97.7}\\ \hline
\multicolumn{1}{c|}{\textbf{CT}}               		& \cellcolor[HTML]{D5E2F1}{7.65} & \textcolor{minired}{98.1}    & \cellcolor[HTML]{D5E2F1}{11.2} & 98.2 & \cellcolor[HTML]{F3CECA}{\textcolor{minired}{90.2}} & 98.4  & \cellcolor[HTML]{D5E2F1}{1.27} & 98.5 &27.6 &98.3 & 90.2 & \multicolumn{1}{|c}{\textbf{98.1}}\\ \hline
\multicolumn{1}{c|}{\textbf{Ours}}              &\cellcolor[HTML]{D5E2F1}{8.93} & \textcolor{minired}{98.1} & \cellcolor[HTML]{D5E2F1}{\textcolor{minired}{15.4}} & 98.1 & \cellcolor[HTML]{D5E2F1}{1.44} & 99.2 & \cellcolor[HTML]{D5E2F1}{0.36} & 98.5 &\textbf{6.53} &\textbf{98.5}& \textbf{15.4} & \multicolumn{1}{|c}{\textbf{98.1}}\\ \hline
\end{tabular}}
\vspace{-1.em}
\caption{
Upstream and Downstream Evaluation and comparison results under \textbf{\underline{Case-2}} with STL-10.
}
\label{tab:TLstl10}
\vspace{-.5em}
\end{table}

\vspace{-1em}
\subsection{Ablation Study}
\label{sec:ablation}
\vspace{-.5em}


Table \ref{tab:differentstru} shows that solely adopting the outer offset loop will experience limitations in low poison ratio cases. 
In the case of a low poison ratio, since the poison samples account for a relatively small proportion in each mini-batch, the model will tend to optimize its output for clean samples, thus ignoring its output for poison samples, finally leading to limited performance.
However, this limitation can be effectively overcome by embedding an inner loop to perform poison concentration. In a relatively high poison ratio setting (e.g., 20\%) where the outer loop alone can already achieve good detection performance, inserting an inner loop is still useful and can further boost the detection efficacy. 
It can be seen that the design of the inner loop is the key to our successful defense in spite of the very low poison ratio in Table \ref{tab:ratioablation}.

\begin{table}[t!]
\centering
\resizebox{0.6\columnwidth}{!}{
\begin{tabular}{c|cc|cc} 
\hline
\multirow{2}{*}{}        & \multicolumn{2}{c|}{\textbf{BadNets (5\%) }} & \multicolumn{2}{c}{\textbf{BadNets (20\%) }}  \\ 
\cline{2-5}
                         & \textbf{TPR}  & \textbf{FPR}                 & \textbf{TPR}  & \textbf{FPR}                  \\ 
\hline
\textbf{Outer loop only} & 39.0          & 37.3                        & 96.6          & 0.83                          \\ 
\hline
\textbf{Outer + Inner}   & \textbf{99.5} & \textbf{5.24}                & \textbf{99.9} & \textbf{0.03}                \\
\hline
\end{tabular}}
\vspace{-1.em}
\caption{
Detection effects w/ or w/o inner loop (\underline{\textbf{Case-0}}).
}
\label{tab:differentstru}
\vspace{-1.5em}
\end{table}

We ablate on the size of the base set used in our detection, and the result is provided in Table \ref{tab:basesetsize}. We find that the detection performance slightly decreases as the base set size is smaller; nevertheless, $\AlgName$ can achieve strong performance even with $10$ samples---one sample per class on CIFAR-10. Our experiment confirms our conclusion in Section \ref{sec:keyidea} that the base set and the clean portion of the poisoned dataset share the same clean distribution, while the clean sample and poison sample originate from distinct distributions.

\begin{figure}[t!]
  \centering
     \vspace{-1.5em}
  \includegraphics[width=0.65\linewidth]{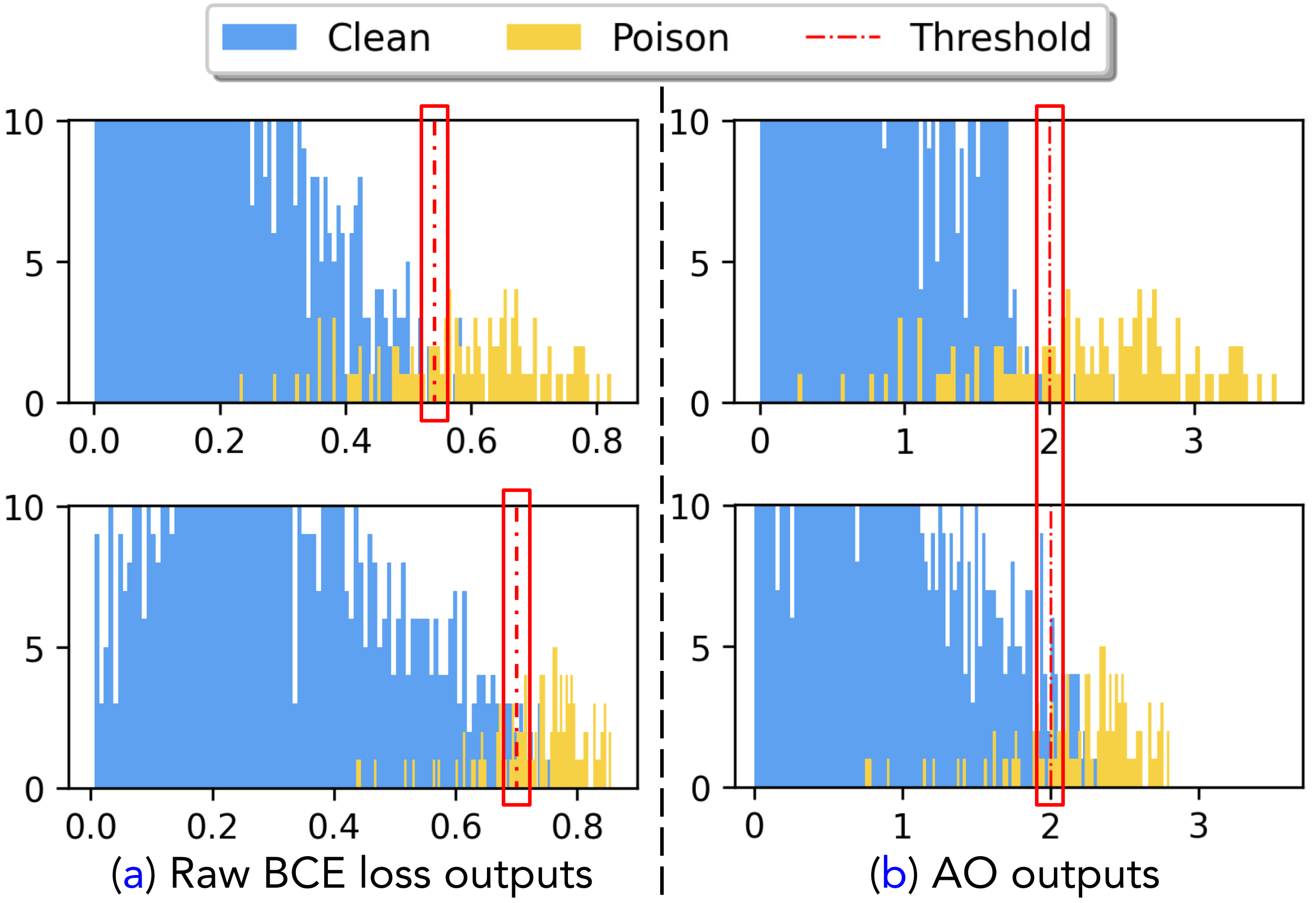}
  \vspace{-1.em}
  \caption{
  The original BCE loss output \textbf{(\textcolor{blue}{a})} and the output processed after AO
\textbf{(\textcolor{blue}{b})} (WaNet attack, CIFAR-10, ResNet-18, \underline{\textbf{Case-0}}). In particular, AO maps the original outputs to a more separable range which is easier to concentrate the poisoned samples with a fixed threshold. 
  }
  \label{fig:mentoutput}
  \vspace{-1em}
\end{figure}

Figure \ref{fig:mentoutput} depicts AO's impact on mini-batches from the same poisoned training set. In particular, even though the two mini-batches are from the same distribution, the number of poisoned samples varies due to random sampling. With different sizes of poisoned samples resulting in different distributions of the loss values, it becomes harder for the inner loop to use a fixed threshold or fixed ratio to determine the most likely poisoned samples to form $B_{\text{pc}}$. AO helps to map the distribution adaptively so that we find a fixed threshold to consistently obtain the poison-concentrated subset.

\begin{table}[h!]
\centering
\resizebox{0.8\columnwidth}{!}{
\begin{tabular}{c|cc|cc|cc|cc}
\hline
\multirow{2}{*}{} &
  \multicolumn{2}{c|}{\textbf{10}} &
  \multicolumn{2}{c|}{\textbf{100}} &
  \multicolumn{2}{c|}{\textbf{1000}} &
  \multicolumn{2}{c}{\textbf{5000}} \\ \cline{2-9} 
                 & TPR & FPR & TPR & FPR  & TPR & FPR & TPR & FPR \\ \hline
\textbf{BadNets (5\%)} & 98.2         & 1.04         & 98.9         & 1.0                 & 99.5         & 0.55         & 99.5         & 0.22         \\ \hline
\textbf{Blended (5\%)} & 100          & 0.18         & 99.9         & 0.01               & 100          & 0.00            & 100          & 0.00            \\ \hline
\end{tabular}}
\vspace{-1.em}
\caption{Ablation study in the base set size (\underline{\textbf{Case-0}}).
}
\label{tab:basesetsize}
\vspace{-1.5em}
\end{table}

\vspace{-1.em}
\subsubsection{Impact of Base Set Quality on Detection Efficacy}
\vspace{-.5em}
While $\AlgName$ exhibits robust performance across a range of attack settings, its effectiveness may fluctuate depending on the quality of the base set.

\begin{table}[h]
\centering
\resizebox{0.95\columnwidth}{!}{
\begin{tabular}{c|cc||cc|cc|cc}
\hline
\multirow{2}{*}{ } &
  \multicolumn{2}{c||}{\textbf{CIFAR-10}} &
  \multicolumn{2}{c|}{\textbf{CIFAR-100}} &
  \multicolumn{2}{c|}{\textbf{STL-10}} &
  \multicolumn{2}{c}{\textbf{GTSRB}} \\ \cline{2-9} 
 &
  TPR $\uparrow$&
  FPR $\downarrow$&
  TPR $\uparrow$&
  FPR $\downarrow$&
  TPR $\uparrow$&
  FPR $\downarrow$&
  TPR $\uparrow$&
  FPR $\downarrow$\\ \hline
\textbf{BadNets (5\%)} &
  99.5 &
  0.55 &
  98.6 &
  0.81 &
  87.3 &
  3.21 &
  0.00 &
  100.0 \\ \hline
\end{tabular}}
\vspace{-1em}
\caption{
Use non-iid dataset as the base set (\underline{\textbf{Case-0}}). The CIFAR-10 column represents the iid setting.
}
\label{tab:iidbase}
\vspace{-1em}
\end{table}

\noindent
\textbf{Sampling quality of the base set.}
In this paper, the base set follows the widely accepted setting \cite{zeng2021adversarial,qi2022fight} that it is drawn from the same distribution as the training set. However, it is worth noting that in practical, a distributional drifts may occur between the training and base sets. To test how $\AlgName$ fares in the face of such distributional drifts, we have outlined the detection results derived from utilizing samples taken from different datasets as base sets for poison detection on CIFAR-10 (BadNets attack, \textbf{\underline{Case-0}}) in Table \ref{tab:iidbase}.
Our observations suggest that $\AlgName$ can consistently generate acceptable detection results if the distributional drift does not drastically alter the task context, as evidenced by the results from CIFAR-100 and STL-10. However, the detection efficiency falters when an out-of-distribution dataset is used as the base set, as exemplified by the use of the traffic sign dataset, GTSRB.

\begin{table}[h!]
\centering
\resizebox{0.95\columnwidth}{!}{
\begin{tabular}{c|cc||cc|cc|cc}
\hline
\multirow{2}{*}{ } &
  \multicolumn{2}{c||}{\textbf{0/1000}} &
  \multicolumn{2}{c|}{\textbf{1/1000}} &
  \multicolumn{2}{c|}{\textbf{5/1000}} &
  \multicolumn{2}{c}{\textbf{10/1000}} \\ \cline{2-9} 
 &
  TPR $\uparrow$&
  FPR $\downarrow$&
  TPR $\uparrow$&
  FPR $\downarrow$&
  TPR $\uparrow$&
  FPR $\downarrow$&
  TPR $\uparrow$&
  FPR $\downarrow$\\ \hline
\textbf{BadNets (5\%)} &
  99.5 &
  0.55 &
  85.4 &
  1.27 &
  78.7 &
  0.00 &
  9.16 &
  3.39 \\ \hline
\end{tabular}}
\vspace{-1em}
\caption{
Number of poisoned samples in the base set (\underline{\textbf{Case-0}}). The 0/1000 column represents the clean base set.
}
\label{tab:baseclean}
\vspace{-1.em}
\end{table}
\noindent
\textbf{Poisons in the base set.}
Stronger attack settings may enable attackers to tamper with the base set. Implementing this setting is challenging, and it has rarely been discussed in prior work due to the formidability of embedding the exact trigger into the carefully scrutinized base set without triggering any alerts. We evaluate the impact of different poison ratios in the base set in Table \ref{tab:baseclean}, and with 10 poisoned samples infiltrating the base set will cause the detection to be ineffective.
%

\noindent
\textbf{Remark.}
The above results on the efficacy and the base set quality are unsurprising. The detection efficacy's sensitivity to the quality of the base set is not exclusive to $\AlgName$. This sensitivity is likewise a noted drawback of numerous defensive methods that rely on a clean in-distribution base set, as observed and discussed in \cite{zeng2022sift}. The experimental results highlight the importance of obtaining high-quality base sets with the care of drift and security inspections. How to effectively acquire a high-quality base set is out of the scope of this paper.

\end{document}